\pdfoutput=1

\documentclass[11pt]{article}

\usepackage[final]{acl}
\usepackage{kotex}
\usepackage{stfloats}
\usepackage{dsfont}

\usepackage{times}
\usepackage{latexsym}

\usepackage{arydshln}

\usepackage[T1]{fontenc}

\usepackage[utf8]{inputenc}

\usepackage{microtype}

\usepackage{inconsolata}

\usepackage{graphicx}
\usepackage{colortbl}
\usepackage{multirow}
\usepackage{arydshln}
\usepackage{booktabs}
\usepackage{amssymb}
\usepackage{amsmath}

\usepackage{float}

\usepackage{algorithm}
\usepackage{algpseudocode}

\usepackage{enumitem}
\usepackage{setspace}
\usepackage{makecell}

\usepackage{thmtools} 
\usepackage{thm-restate}
\usepackage{amsthm}

\usepackage[dvipsnames, table]{xcolor}
\usepackage{pifont}

\definecolor{ssoyblue}{rgb}{0.01, 0.28, 0.9}

\newcommand{\ours}{Q-TIE}
\newcommand{\tempint}{$\langle t_{\text{start}}, t_{\text{end}} \rangle$}
\newcommand{\tscore}{$S_\text{temp}$}
\newcommand{\sscore}{$S_\text{sem}$}
\definecolor{ForestGreen}{RGB}{0,100,0}
\newcommand{\red}[1]{\textcolor{red}{#1}}
\newcommand{\green}[1]{\textcolor{ForestGreen}{#1}}
\newcommand{\cmark}{{\color{green!60!black}\checkmark}}
\newcommand{\xmark}{{\color{red!70!black}\ding{55}}}

\usepackage{todonotes}
\usepackage{xcolor}

\title{\ours{}: A Lightweight and Generalizable Re-ranking Framework for Temporal Information Retrieval}

\author{
Soyeon Kim$^{1}$,
Hyunjin Kim$^{2}$,
JinYeong Bak$^{2}$,
Steven Euijong Whang$^{1\thanks{Corresponding author}}$
\\
$^{1}$Korea Advanced Institute of Science and Technology
\\
$^{2}$Sungkyunkwan University
\\
\texttt{\{purplehibird, swhang\}@kaist.ac.kr}
\\
\texttt{\{khyunjin1993@g.skku.edu, jy.bak@skku.edu\}}
}

\begin{document}
\nocite{*}
\maketitle 

\begin{abstract}
Temporal Information Retrieval (TIR) has been increasingly critical given the rise of Retrieval-Augmented Generation (RAG). Since temporally mismatched evidence can be highly misleading, TIR aims to retrieve documents that are both semantically and temporally relevant to a query. Two TIR paradigms have emerged -- temporal retrievers and temporal re-rankers -- differing in how temporal relevance is modeled. While these paradigms provide complementary strengths, our analysis reveals that each alone falls short of robust TIR: temporal retrievers provide flexible query understanding via learned representations, but often fail to explicitly account for temporal constraints; temporal re-rankers can enforce such constraints more explicitly, but often rely on predefined re-ranking rules. To address this, we propose \textbf{\ours{}}, a re-ranking framework based on learned Temporal Intent Extraction (TIE). By introducing a TIE model that maps each query's temporal constraint into a unified interval representation (i.e., \tempint{}), \ours{} generalizes beyond predefined rules via model-based learning while explicitly modeling temporal constraints as a separate signal -- jointly achieving what each paradigm typically trades off. Experiments demonstrate that \ours{} consistently outperforms existing TIR methods with stronger generalizability across temporal query types, and provides a lightweight yet effective add-on for temporally-aware RAG pipelines. Code: \url{https://github.com/ssoy0701/Q-TIE}.
\end{abstract}

\section{Introduction}
As the world evolves, the correctness of information changes over time -- making \textit{temporal awareness} essential for retrieval systems. Even a topically relevant document can be highly misleading if it corresponds to the wrong time period. For example, in Retrieval-Augmented Generation (RAG) scenarios~\citep{lewis2020retrieval}, temporally mismatched evidence has been shown to cause Large Language Models (LLMs) to generate outdated or time-inconsistent answers~\citep{kim2025harnessing, han2025rag}. These risks have drawn attention to Temporal Information Retrieval (TIR)~\citep{campos2014survey, piryani2025s}, which seeks to retrieve documents that are both relevant to the query topic (i.e., semantic relevance) and the timeframe requested by the query (i.e., temporal relevance).

\vspace{-0.2cm}
\begin{figure*}
    \centering
    \includegraphics[width=0.99\textwidth]{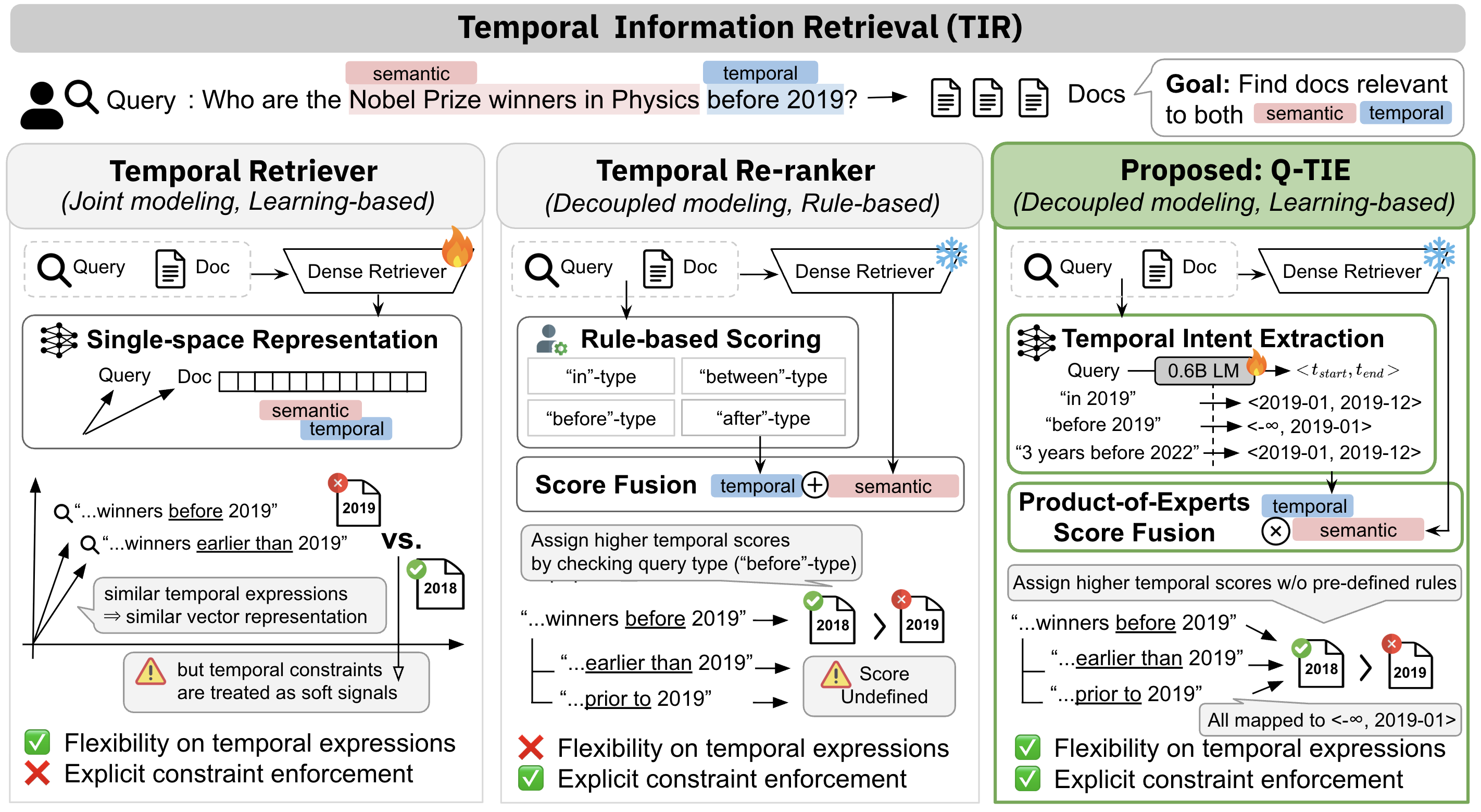}
    \vspace{-0.35cm}
\caption{\textbf{Comparison with prior TIR approaches.} Temporal retrievers (left) jointly encode semantic and temporal relevance in a single space and offer flexibility for diverse temporal expressions, but struggle to enforce hard temporal constraints. Temporal re-rankers (middle) decouple temporal scoring via rule-based query classification and strictly enforce temporal constraints, but struggle on unseen expressions. Our \ours{} (right) resolves this trade-off via a decoupled yet learning-based design: a 0.6B LM maps arbitrary temporal expressions into explicit intervals for temporal scoring, aggregated with semantic scoring via Product-of-Experts. By this design, \ours{} offers both flexibility on temporal expressions via learned mapping and explicit constraint enforcement via decoupled scoring.}
    \label{fig:ours_framework}
\vspace{-0.2cm}
\end{figure*}

\smallskip \smallskip
Two paradigms have emerged for TIR, differing in how temporal relevance is modeled: jointly with semantic relevance
or separately from it. The first, \textit{temporal retrievers}, jointly learn semantic and temporal relevance by fine-tuning dense retrievers on temporally supervised data~\citep{wu2024time, abdallah2025tempretriever, han2025temporal}. The second, \textit{temporal re-rankers}, model temporal relevance as a separate signal while keeping the dense retriever fixed~\citep{siyue2024mrag, qian2024timer4, gade2025s}. These methods re-rank documents using additional temporal relevance scores, often computed via pre-defined rules per query type (e.g., ``before''-type, ``in''-type).

Despite progress, our analysis reveals complementary gaps in both paradigms -- each failing where the other succeeds -- which limit generalizability across 
diverse temporal queries. Temporal retrievers can flexibly capture diverse 
temporal expressions by jointly learning semantic and temporal relevance 
directly from data, without relying on pre-defined rules. However, this joint 
learning reduces temporal constraints to an implicit soft signal, causing them to frequently rank temporally invalid documents 
highly. For example, for a ``before 2019'' query, documents published after 
2019 should be filtered out, yet temporal retrievers often retrieve them as 
top results (see Fig.~\ref{fig:tir_per_query}). Temporal re-rankers, by 
contrast, can enforce constraints more strictly by decoupling temporal relevance from semantic relevance; however, they often rely on pre-defined 
keyword-matching rules, causing a 33.1\%p Precision@1 drop on out-of-template 
queries. Together, these findings suggest that neither paradigm jointly 
achieves the two properties essential for robust and generalizable TIR: 
strong temporal constraint enforcement and flexibility beyond pre-defined rules.

To address this limitation, we propose \textbf{\ours{}}, a new temporal re-ranking framework with \textbf{Q}uery-based \textbf{T}emporal \textbf{I}ntent \textbf{E}xtraction. Instead of relying on predefined rules, \ours{} introduces a learned task -- \textit{Temporal Intent Extraction} (TIE) -- that maps each timeframe required in a query to a unified representation, \tempint{}. As illustrated in Fig.~\ref{fig:ours_framework}, this learned task allows \ours{} to flexibly recognize diverse temporal expressions beyond fixed rules, while explicitly enforcing temporal constraints by separating the temporal signal. To train a dedicated TIE model, we newly construct a dataset covering 13 temporal query types with diverse linguistic expressions. The trained TIE model extracts temporal intent from queries to compute temporal relevance scores, which is then integrated with semantic relevance scores via a Product-of-Experts (PoE) formulation~\citep{hinton2002training}. Unlike 
standard sum-based aggregation~\citep{gade2025s}, PoE requires both scores 
to be high through multiplicative fusion, effectively preventing high 
semantic similarity from overshadowing temporal invalidity.

We also design \ours{} as a lightweight framework to enable practical RAG deployments. The TIE model is built on a 0.6B-parameter language model, yet achieving TIE performance comparable to much larger models (e.g., Gemma3-12B) and general neural re-rankers (e.g., Qwen3 re-rankers) at substantially lower inference cost. This efficiency makes \ours{} a cost-effective alternative for modern RAG pipelines,  which often incur multiple retrieval calls for improved RAG performances~\citep{asai2024self, yan2024corrective}.

Experiments show that \ours{} substantially reduces performance gaps across diverse temporal query types, outperforming both temporal retrievers and temporal re-rankers. Further analysis demonstrates \ours{}'s compatibility with existing RAG strategies such as query rewriting~\citep{qian2024timer4}, showing its potential to extend coverage to more complex temporal queries.

\smallskip
\noindent\textbf{Contributions.\quad} Our primary contributions are:
\begin{itemize}[leftmargin=0.5cm]
\vspace{-0.25cm}
\setlength\itemsep{-1pt}
    \item We identify generalizability across temporal query types as an overlooked challenge in TIR, and show that existing paradigms exhibit complementary failure modes.

    \item We propose \ours{}, a new temporal re-ranking framework that achieves both explicit temporal constraint enforcement and out-of-template generalization, two properties that prior methods have typically traded-off.

    \item We empirically show that \ours{} consistently outperforms existing TIR methods across diverse temporal query types, and serves as a lightweight yet effective drop-in for temporally-aware RAG pipelines.

\end{itemize}

\section{Related Work}
\label{sec:related_work}

\paragraph{Temporal Information Retrieval.} 
Temporal Information Retrieval (TIR) aims to retrieve documents
relevant to a query in both topic (i.e., semantic
relevance) and timeframe (i.e., temporal
relevance)~\citep{campos2014survey, piryani2025s}. One line
of work, \textit{temporal retrievers}, trains end-to-end
retrievers on temporally supervised data to jointly encode
semantic and temporal relevance into a single representational
space~\citep{wu2024time, abdallah2025tempretriever,
han2025temporal}. Another line, \textit{temporal re-rankers},
keeps the base retriever fixed and applies a separate temporal
scoring step; because different query types require different
scoring functions (e.g., ``before''- vs. ``after''-type queries require opposite scoring of documents),
these methods identify the query's temporal type via
keyword-matching rules and apply the corresponding predefined
scoring function~\citep{siyue2024mrag, gade2025s,
qian2024timer4}. As we illustrate in
Fig.~\ref{fig:ours_framework}, temporal retrievers frequently
surface temporally invalid documents due to soft temporal
constraints, while temporal re-rankers fail to generalize
beyond predefined templates. \ours{} bridges
both gaps by learning to extract temporal intent,
flexibly handling diverse expressions and 
explicitly enforcing temporal constraints in a
retriever-agnostic re-ranking formulation.

\paragraph{Temporal Expression Understanding.}

Our proposed Temporal Intent Extraction (TIE) task -- a learned task to extract temporal intent from queries -- is rooted in temporal expression understanding in NLP. Early work has taken a rule-based approach~\citep{strotgen2010heideltime, chang2012sutime, bethard2013synchronous}, looking up tokens in a normalization lexicon and mapping them to a specific format. Neural approaches have since improved robustness over diverse surface forms of temporal expressions~\citep{laparra2018characters, lange2023multilingual}, but remain limited by domain dependency, struggling to generalize across domains without retraining. Temporal expression understanding has also been explored in time-sensitive question answering scenarios~\citep{chen2021dataset, tan2023towards}, yet the goal there is direct answer generation. In contrast, TIE is specialized for retrieval as the end task: instead of standardizing surface expressions or generating final answers, the task is to \textit{infer} the temporal interval that a relevant document must satisfy -- a formulation supported by a newly constructed dataset covering the full range of Allen's temporal relations~\citep{allen1983maintaining} between queries and documents in TIR settings.

\paragraph{Temporally-aware RAG.}
Retrieval-Augmented Generation (RAG) enhances the quality of
LLM-generated answers by grounding them in externally
retrieved evidence~\citep{lewis2020retrieval, gao2023retrieval};
however, recent studies have shown that temporally invalid evidence often cause LLMs to produce
outdated or time-inconsistent answers~\citep{kim2025harnessing, han2025rag}, necessitating temporal
awareness in RAG. To address this, existing RAG frameworks primarily employ temporal
retrievers or re-rankers at the retrieval stage~\citep{piryani2025s}, where re-rankers are often preferred because their retriever-agnostic nature avoids additional retriever retraining~\citep{siyue2024mrag, gade2025s}.
\ours{} falls into this category, but goes further by replacing
rule-based temporal scoring with a learning-based approach
for greater flexibility over diverse temporal expressions.
Compared to general neural re-rankers~\citep{zhang2025qwen3} and prompt-based LLM re-rankers~\citep{chen2026defuzzrag}, \ours{} provides
temporal specialization at substantially lower cost (0.6B
parameters), making it a practical drop-in to achieve
temporal awareness for modern RAG pipelines that often
incur multiple retrieval calls to improve performance~\citep{asai2024self, yan2024corrective}.

\begin{table*}[h]
\centering
\small
\begin{tabular}{l@{\hspace{3pt}}c@{\hspace{6pt}}p{13cm}@{}}
\toprule
\textbf{Type} & \textbf{Count} & \textbf{Example Query} \\
\midrule \midrule
before        & 396 & Who were the Presidents of Pakistan whose term ended before July 31, 2018? \\
met-by        & 261 & Which individual held the role of President in Brazil beginning two months prior to May 15, 1990? \\
meet          & 265 & Can you identify the team Fernando Alonso raced for in F1 three years prior to January 1, 2009? \\
start         & 150 & Who started their term as CEO in 2017? \\
started-by    & 150 & Who was the CEO of Walmart in 1988 and held the position until before 2004? \\
finish        & 150 & Which team did Neymar Jr. play for with a contract starting after Dec.\ 31, 2009, and ending within 2017? \\
equal         & 150 & Which people held the role of director at the IAEA during the years 1957 to 1961? \\
finished-by   & 150 & Which South Korean President's term concluded in February 1993? \\
overlap       & 150 & Who served as President of Italy and started their term before Jan.\ 31, 2013, ending it between Jan.\ 31, 2013, and Jan.\ 14, 2018? \\
overlapped-by & 150 & Which individual held the role of President in Indonesia, beginning their term between Jul.\ 23, 1999, and Feb.\ 28, 2003, and remained in office after Feb.\ 28, 2003? \\
contain       & 129 & What soccer team was Harry Kane contracted to before Jan.\ 1, 2012, and after Dec.\ 31, 2013? \\
after         &  100 & Who won the Nobel Prize in Economics later than 2020? \\
during        &  100 & Can you list the names of individuals who held the role of general secretary at NATO from 1971 to 1984? \\
\bottomrule
\end{tabular}
\vspace{-0.2cm}
\caption{\textbf{Query distribution across 13 Allen relations in newly constructed TIE dataset.} We show sample counts and example queries for each type. The total of 2,300 queries are split into train and test sets with an 80/20 ratio.}
\label{tbl:tie_stats}
\vspace{-0.3cm}
\end{table*}

\begin{table}[]
    \small
    \centering
    \begin{tabular}{p{0.95\linewidth}}
    \toprule
    \textbf{TIE Task Format} \\
    \midrule\midrule
    \texttt{Extract the time span from the question.}\\
    \texttt{Question: \textless query\textgreater}\\
    \texttt{Answer:} \\     
    \texttt{start: \textless YYYY-MM or None\textgreater}\\
    \texttt{end: ~~\textless YYYY-MM or None\textgreater}\\
    \bottomrule
    \end{tabular}
    \vspace{-0.2cm}
    \caption{\textbf{Proposed TIE task format.} The exact training/inference prompts are presented in \S\ref{supp:tie_training}.}
    \label{tbl:tie_task_format}
\vspace{-0.2cm}
\end{table}

\section{\ours{} Framework}
\label{sec:ours}

\paragraph{Problem Formulation.}
We formulate TIR as a document re-ranking task over two
criteria: \textit{semantic relevance} and \textit{temporal
validity}. Given a temporal query $q$, a base retriever
returns candidate documents $\mathcal{D}= \{d_1, \ldots,
d_n\}$ ranked by its semantic relevance score
$S_\text{sem}(q, d_i)$, where each $d_i$ is associated with
a timestamp $\tau_i$. Our goal is to re-rank $\mathcal{D}$
such that top-ranked documents are both semantically relevant
and temporally valid to the timeframe expressed
in $q$.

\paragraph{Challenges.}
We identify three challenges in TIR:
\begin{itemize}[leftmargin=0.5cm]
\vspace{-0.25cm}
\setlength\itemsep{-1pt}
\item \textbf{(C1) Linguistic variation in temporal queries.}
The same temporal constraint can take widely varied forms (e.g., ``before
2019'', ``prior to 2019'', ``no later than 2019''), where a reliable TIR
system must interpret them consistently.

\item \textbf{(C2) Complementary gaps in temporal relevance modeling.}
As illustrated in Fig.~\ref{fig:ours_framework}, existing TIR approaches often
face a trade-off in how they model temporal relevance: temporal retrievers
offer flexibility via learned representations, but treat temporal constraints
as soft signals; rule-based re-rankers strictly enforce constraints, but are
bound to pre-defined templates. Yet, flexibility on temporal expressions and
explicit constraint enforcement are both essential for robust TIR.

\item \textbf{(C3) Aggregating semantic and temporal relevance.}
Even with an accurate temporal relevance score, TIR requires both signals to
hold jointly. Naive aggregation can let one signal overwhelming the other, leading to undesired retrieval (e.g., surfacing semantically similar yet temporally invalid documents at the
top).
\end{itemize}

\paragraph{Roadmap.} 
\ours{} addresses these challenges as follows. We first introduce the Temporal
Intent Extraction (TIE) task (\S\ref{subsec:tie}), which tackles C1 via a newly
constructed dataset covering diverse
temporal query types with varied surface expressions. For C2, \ours{} takes a hybrid approach that combines the strengths of both
paradigms: decoupled temporal scoring (\S\ref{subsec:temporal_scoring})
preserves explicit constraint enforcement, while a learned TIE model provides
flexibility over diverse expressions.
Finally, the temporal score $S_{\text{temp}}$ is aggregated with
$S_{\text{sem}}$ via a Product-of-Experts formulation (\S\ref{subsec:poe_aggregation}),
addressing C3 to yield the balanced final ranking score $S_{\text{final}}$.

\subsection{Temporal Intent Extraction (TIE)}
\label{subsec:tie}

We define TIE as the task of mapping a query $q$ to a temporal 
window $\langle t_{\text{start}}, t_{\text{end}} \rangle$ 
representing the query's target time range (Table~\ref{tbl:tie_task_format}). The core of this task is to 
reduce diverse linguistic expressions to a unified interval 
representation, enabling both consistent handling of varied 
forms and explicit identification of temporal constraints. We use 
\texttt{None} to express open-ended ranges (e.g., ``before 
2019'' as $\langle \texttt{None}, \texttt{2019-01} \rangle$) 
and month-level granularity for finer extraction than the year-level granularity typically adopted in TIR~\citep{piryani2025s}.

\subsubsection{TIE Dataset Construction}
\label{subsubsec:tie_data}

To support the TIE task, we construct a dataset covering a 
comprehensive set of temporal query types with diverse 
linguistic expressions; dataset statistics are summarized in 
Table~\ref{tbl:tie_stats}.

\paragraph{Temporal Query Types.} 
To ensure comprehensive coverage of temporal relations, we 
ground our query types in Allen's interval 
algebra~\citep{allen1983maintaining}, which defines 13 
mutually exclusive and exhaustive relations between two time 
intervals: \textit{before}, \textit{after}, \textit{meet}, 
\textit{met-by}, \textit{overlap}, \textit{overlapped-by}, 
\textit{start}, \textit{started-by}, \textit{during}, 
\textit{contain}, \textit{finish}, \textit{finished-by}, and 
\textit{equal}. Whereas prior work handles only 4--6 relation
types~\citep{qian2024timer4, siyue2024mrag}, grounding in this
well-established taxonomy extends our coverage to all 13 relations.

\paragraph{Data Construction Process.} 
We build on two complementary benchmarks: one that supplies 
training signal across all 13 Allen relations, and one that 
provides a corpus for end-to-end TIR evaluation. 
\textit{TDBench}~\citep{kim2025harnessing} covers all 13 
relations and pairs each query with SQL constraints that 
explicitly encode its temporal window; we parse these 
constraints with a rule-based script to recover ground-truth 
$\langle t_{\text{start}}, t_{\text{end}} \rangle$ labels. 
\textit{Nobel Prize}~\citep{wu2024time} contributes a corpus 
with challenging hard negatives -- documents that are 
semantically relevant yet temporally mismatched -- making it 
well-suited for evaluating downstream TIR performance after the TIE task. As its original queries cover only basic expressions (e.g., \textit{in},
\textit{between}), we manually curate additional queries with complex
constraints (e.g., \textit{3 years before}) and annotate their temporal
windows. Full dataset construction details are provided in \S\ref{supp:tie_data_process}.

\subsubsection{TIE Model Training}
\label{subsubsec:tie_model}
Building on the constructed dataset, we train a dedicated TIE model that
learns to map diverse temporal queries to their temporal windows (Table~\ref{tbl:tie_task_format}).
We instantiate the TIE model by fine-tuning 
Qwen3-0.6B~\citep{yang2025qwen3}, chosen to balance 
extraction accuracy with inference efficiency. We train on 
the constructed dataset with 
Low-Rank Adaptation (LoRA)~\citep{hu2022lora} for parameter-efficient training; see training prompts and hyperparameters in \S\ref{supp:tie_training}. Despite its compact size (0.6B), experiments in \S\ref{sec:experiments} demonstrate that the trained TIE model achieves comparable TIE performance to much larger 7B--9B 
models, confirming the 
effectiveness of our targeted supervision with the constructed dataset. Ablations across model families and sizes
(\S\ref{supp:exp_tie_model_ablation}) show that scaling the backbone up to 8B
yields only marginal gains, supporting the 0.6B backbone as a favorable
cost-performance choice.

\subsection{Temporal Relevance Scoring}
\label{subsec:temporal_scoring}
\vspace{-0.1cm}

Given the query's temporal window $\langle t_{\text{start}},
t_{\text{end}} \rangle$ extracted by the TIE model and a document's timestamp
$\tau_i$, we compute a temporal relevance score \tscore{} reflecting how well
the document falls within the queried timeframe. A natural first choice is
strict interval matching, assigning 1 if the document time overlaps the query
window and 0 otherwise. This, however, scores a near-miss document
identically to a far-off one, discarding a useful ranking signal between them. We thus
compute a \textit{soft overlap} via a Gaussian kernel:
\setlength{\abovedisplayskip}{4pt}
\setlength{\belowdisplayskip}{4pt}
\begin{equation}
\scalebox{0.94}{$
S_{\text{temp}}(q, d) = \sum_{i,j} p_{q,i}\, p_{d,j}\,
\exp\!\left(-\frac{(t_i - t_j)^2}{2\sigma^2}\right)
\Delta t_i \Delta t_j
$,}
\label{eq:time_score}
\end{equation}
where $p_q$ and $p_d$ are uniform distributions over the query window and the
document's timestamp, and $i, j$ index discretized time points with widths
$\Delta t_i, \Delta t_j$. Here, the Gaussian kernel scores each document by
its temporal closeness to the query window: the score peaks when the document
falls inside the window and decays smoothly as it moves away. The decay rate
is controlled by $\sigma$, with smaller values approaching strict interval
matching. This soft scoring prioritizes documents within the queried timeframe, while
still retaining fine-grained ranking signals for documents outside it.

\subsection{Score Aggregation via Product-of-Experts}
\label{subsec:poe_aggregation}

\paragraph{Aggregation.}
To produce the final ranking, we combine the semantic score 
\sscore{} from the base retriever with the temporal score 
\tscore{} via a log-sum 
fusion, which leads to a Product-of-Experts (PoE) 
formulation~\citep{hinton2002training}: 
\begin{equation}
\scalebox{0.94}{$
\begin{aligned}
S_{\text{final}}(q, d) &= S_{\text{sem}}(q, d) + \lambda \cdot 
\log S_{\text{temp}}(q, d) \\
&\approx \log\!\left(P_{\text{sem}} \cdot 
P_{\text{temp}}^{\lambda}\right) + C,
\end{aligned}
$}
\label{equ:poe_fusion}
\end{equation}
where $\lambda$ controls the relative weight of the temporal 
score. The
theoretical basis of this form follows energy-based
models~\citep{lecun2006tutorial}, interpreting each score as an unnormalized
log-probability from two experts: a semantic expert and a temporal expert. The log-sum fusion turns score addition in Eq.~\ref{equ:poe_fusion}
into multiplication in probability space (i.e., $P_{\text{sem}} \cdot
P_{\text{temp}}^{\lambda}$), up to a normalization constant $C$. A detailed
derivation is provided in \S\ref{supp:poe_formulation}.

\paragraph{Effect.}
The multiplicative form in Eq.~\ref{equ:poe_fusion} imposes an \textit{AND-like}
aggregation: a document ranks highly only when both experts agree, since a
near-zero \tscore{} sharply penalizes
the final score with a large negative log term. In contrast, standard sum-based fusion in prior
TIR~\citep{gade2025s} behaves \textit{OR-like}: a high semantic score alone can
override a clear temporal mismatch, allowing temporally invalid documents to
rank highly. In \S\ref{sec:experiments}, we empirically confirm this contrast and the advantage of PoE for
TIR, where both signals must hold.

\begin{table*}[t]
\centering
\scalebox{0.69}{
\begin{tabular}{@{}l|r@{\hspace{8pt}}r@{\hspace{8pt}}r@{\hspace{8pt}}r| r@{\hspace{8pt}}r@{\hspace{8pt}}r@{\hspace{8pt}}r| r@{\hspace{8pt}}r@{\hspace{8pt}}r@{\hspace{8pt}}r| r@{\hspace{8pt}}r@{\hspace{8pt}}r@{\hspace{8pt}}r@{\hspace{4pt}}|}
\toprule
 & \multicolumn{8}{c|}{\textbf{Nobel Prize}} & \multicolumn{8}{c|}{\textbf{TempRAGEval}} \\
 & \multicolumn{8}{c|}{Simple\hspace{100pt}Complex}
 & \multicolumn{8}{c|}{Simple\hspace{100pt}Complex} \\
\cmidrule(lr){2-5}\cmidrule(lr){6-9}
\cmidrule(lr){10-13}\cmidrule(lr){14-17}
\textbf{Method}
& P@1 & P@5 & R@1 & R@5
& P@1 & P@5 & R@1 & R@5
& P@1 & P@5 & R@1 & R@5
& P@1 & P@5 & R@1 & R@5 \\
\midrule \midrule
\multicolumn{17}{c}{\textit{Non-temporal and Temporal Retrievers}} \\
\midrule
BM25
& 15.60 & 14.33 & 6.54 & 25.75
& 2.88 & 2.26 & 0.57 & 3.48
& 61.79 & 16.51 & 61.79 & 82.55
& 45.54 & 13.66 & 45.54 & 68.30 \\
Contriever
& 17.69 & 13.14 & 4.72 & 16.89
& 10.32 & 10.32 & 2.59 & 12.27
& 61.32 & 17.55 & 61.32 & 87.74
& 55.95 & 16.46 & 55.95 & 82.29 \\
JinaAI
& 38.47 & 24.68 & 12.77 & 35.58
& 15.87 & 13.46 & 5.30 & 22.03
& 65.57 & 17.64 & 65.57 & 88.21
& 60.42 & 16.99 & 60.42 & 84.97 \\
\midrule
TempRetriever
& 18.28 & 10.47 & 5.42 & 15.49
& 7.69 & 5.48 & 1.39 & 8.77
& 54.72 & 15.00 & 54.72 & 75.00
& 42.86 & 12.62 & 42.86 & 63.10 \\
TsContriever
& 58.63 & 36.52 & 19.92 & 52.80
& 13.22 & 14.86 & 4.33 & 27.39
& 77.36 & 18.96 & 77.36 & 94.81
& 46.43 & 16.07 & 46.43 & 80.36 \\
TSM
& 38.13 & 27.17 & 10.69 & 35.73
& 18.75 & 17.45 & 6.00 & 25.75
& 73.11 & 18.87 & 73.11 & 94.34
& 57.89 & 17.14 & 57.89 & 85.71 \\
\midrule
\multicolumn{17}{c}{\textit{Temporal Re-rankers}} \\
\midrule
TempRALM \small(+ Contriever)
& 0.43 & 0.44 & 0.10 & 0.59
& 25.24 & 17.93 & 7.56 & 22.00
& 5.66 & 2.08 & 5.66 & 10.38
& 26.04 & 8.51 & 26.04 & 42.56 \\
MRAG \small(+ Contriever)
& 51.82 & 34.17 & 13.90 & 43.74
& 14.66 & 17.31 & 3.77 & 20.30
& 62.26 & 16.23 & 62.26 & 81.13
& 47.02 & 13.75 & 47.02 & 68.75 \\
TimeR4 \small(+ Contriever)
& 17.69 & 13.14 & 4.72 & 16.89
& 17.31 & 15.00 & 5.82 & 19.73
& 59.91 & 16.98 & 59.91 & 84.91
& 48.96 & 15.33 & 48.96 & 76.64 \\
\midrule
TempRALM \small(+ JinaAI)
& 0.71 & 0.56 & 0.25 & 0.85
& \underline{32.45} & 22.84 & \underline{11.94} & 31.32
& 8.02 & 2.08 & 8.02 & 10.38
& 34.08 & 8.72 & 34.08 & 43.60 \\
MRAG \small(+ JinaAI)
& \underline{66.89} & 38.10 & \underline{20.65} & 53.03
& 18.03 & 18.08 & 4.67 & 25.59
& 63.68 & 15.66 & 63.68 & 78.30
& 32.89 & 9.88 & 32.89 & 49.40 \\
TimeR4 \small(+ JinaAI)
& 38.47 & 24.68 & 12.77 & 35.58
& 21.39 & 18.22 & 8.65 & 30.04
& 65.57 & 16.98 & 65.57 & 84.91
& 57.74 & 16.01 & 57.74 & 80.06 \\
\midrule
\rowcolor{gray!15}
\ours{} \small(+ Contriever)
& 60.54 & \underline{45.25} & 17.96 & \underline{58.91}
& 31.25 & \underline{29.18} & 11.83 & \underline{47.61}
& \underline{88.21} & \underline{19.25} & \underline{88.21} & \underline{96.23}
& \underline{79.32} & \underline{17.68} & \underline{79.32} & \underline{88.39} \\
\rowcolor{gray!15}
\ours{} \small(+ JinaAI)
& \textbf{78.76} & \textbf{55.96} & \textbf{25.47} & \textbf{73.41}
& \textbf{46.88} & \textbf{35.82} & \textbf{18.35} & \textbf{62.30}
& \textbf{93.87} & \textbf{19.53} & \textbf{93.87} & \textbf{97.64}
& \textbf{84.08} & \textbf{18.12} & \textbf{84.08} & \textbf{90.62} \\
\bottomrule
\end{tabular}}
\vspace{-0.2cm}
\caption{\textbf{Overall TIR performance.} We report Precision@$\{1,5\}$ and Recall@$\{1,5\}$
on the simple and complex subsets of Nobel Prize and TempRAGEval datasets. We compare
non-temporal retrievers (BM25, Contriever, JinaAI), temporal
retrievers (TempRetriever, TsContriever, TSM), and temporal
re-rankers (TempRALM, MRAG, TimeR4), where re-rankers and
\ours{} re-rank the top-$k$ documents retrieved by Contriever or
JinaAI. \textbf{Bold} and \underline{underline}
denote the best and second-best results in each column. NDCG
metrics are reported in Table~\ref{tbl:ndcg_retrieval}. Additional results pairing temporal re-rankers with the best performing temporal retriever are reported in Table~\ref{tbl:more_tir}.} 
\label{tbl:nobel_timeqa_retrieval}
\vspace{-0.2cm}
\end{table*}

\section{Experiments}
\label{sec:experiments}

\subsection{Experimental Setup}
\label{subsec:exp_setup}

\paragraph{Datasets.} 
We evaluate \ours{} on two complementary TIR benchmarks, along with our newly constructed dataset. \textit{Nobel Prize}~\citep{wu2024time} offers a
temporally challenging TIR setup: although topically limited to
six Nobel categories, its 989 documents span over a century
(1900--2022), paired with 3,244 queries. In contrast,
\textit{TempRAGEval}~\citep{siyue2024mrag} is semantically
challenging, covering diverse Wikipedia-based topics
across 9,695 documents and 883 queries. To further analyze generalization
across queries, we partition these datasets into \emph{simple}
and \emph{complex} subsets following the TempRAGEval criterion: a
query is simple if its temporal expression is explicit with a
direct year mention that also appears in the document (e.g., ``in 2019''), and
complex otherwise (e.g., ``3 years before'', ``as of''). Since Nobel Prize only contains simple queries (e.g., ``in'',
``between''), we use the augmented held-out test set (\S\ref{subsubsec:tie_data}) as its complex subset; for TempRAGEval, we directly adopt its existing partition, which serves as an out-of-distribution (OOD) test set. More dataset
details are presented in \S\ref{supp:exp_setup_data_stat}.

\paragraph{Baselines.}
We compare \ours{} against five categories of baselines.
(1) \textit{Non-temporal retrievers} (BM25~\citep{robertson2009probabilistic}, Contriever~\citep{izacard2021unsupervised}, JinaAI~\citep{gunther2023jina}) measure semantic relevance without temporal awareness. 
(2) \textit{Temporal retrievers} (TsContriever~\citep{wu2024time}, TempRetriever~\citep{abdallah2025tempretriever}, TSM~\citep{han2025temporal})
jointly model semantic and temporal relevance in a single retriever. (3) \textit{Temporal re-rankers} (TempRALM~\citep{gade2025s}, MRAG~\citep{siyue2024mrag}, TimeR4~\citep{qian2024timer4}) apply 
pre-defined temporal scoring on top of a fixed retriever; for 
MRAG and TimeR4, we isolate their temporal scoring modules from the full RAG pipeline for a head-to-head comparison at the re-ranking level.
(4) \textit{General neural re-rankers} capture
state-of-the-art semantic relevance, but lack temporal
specialization. For the TIE task, we also compare against
(5) \textit{general-purpose LLMs} of varying sizes (0.5B--32B) and types. All temporal baselines are the original fine-tuned versions released by their authors. More baseline details are in
\S\ref{supp:exp_setup_baselines}.

\paragraph{Metrics.}
We report Precision@$k$, Recall@$k$, NDCG@$k$, and F1 score, which are standard metrics for TIR and RAG evaluation~\citep{robertson2009probabilistic}.
Higher values indicate better performance. Formal definitions are presented in \S\ref{supp:exp_setup_metrics}.

\subsection{TIR Performance}
\label{subsec:exp_tir_performance}

\paragraph{Overall Performance.}
Table~\ref{tbl:nobel_timeqa_retrieval} reports retrieval
performance on Nobel Prize and TempRAGEval, split by query
complexity. \ours{} consistently outperforms all baselines by
large margins; for example, on Nobel Prize simple queries,
\ours{} surpasses the strongest temporal retriever (TsContriever)
and re-ranker (MRAG+JinaAI) by 20.13\%p and 11.87\%p Precision@1, respectively. Moreover, baselines degrade sharply on complex queries
(e.g., on TempRAGEval, TsContriever's Recall@1:
77.36$\rightarrow$46.43) while \ours{} remains robust
(93.87$\rightarrow$84.08), demonstrating much
stronger generalization across query complexity levels.

\begin{figure}[t!]
    \vspace{-0.2cm}
    \centering
    \includegraphics[width=0.95\linewidth]{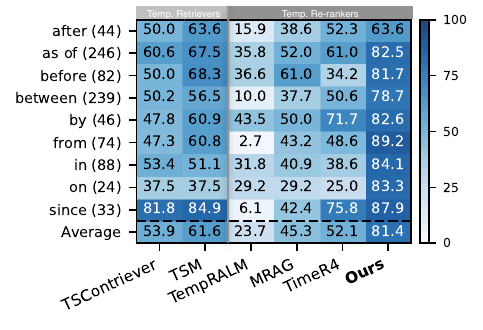}
    \vspace{-0.35cm}
    \caption{\textbf{Per-query-type retrieval performance on
    TempRAGEval, an OOD benchmark for our TIE model.} Precision@1 across nine temporal expression types
    (query counts in parentheses). Re-rankers and \ours{} use Contriever
    as the underlying retriever.}
    \label{fig:tir_per_query}
\end{figure}

\paragraph{Per-Query-Type Performance.}
Fig.~\ref{fig:tir_per_query} further breaks down Precision@1
across nine temporal expression types on TempRAGEval, revealing
a clear tradeoff between existing paradigms. Rule-based
re-rankers excel on types covered by their rules, but fail
elsewhere: TempRALM, whose rule prioritizes recent documents,
scores 35.8/43.5 on \emph{as of}/\emph{by}, but only 2.7/6.1 on
\emph{from}/\emph{since}. Temporal retrievers generalize more
broadly via learning-based scoring, yet can lose to
well-defined rules on covered types (e.g., on \emph{by}, the temporal retriever TSM scores 60.9
while TimeR4's re-ranking reaches 71.7). \ours{} resolves this tradeoff by employing
decoupled yet learning-based design, outperforming temporal retrievers while
generalizing across all types more uniformly than temporal re-rankers.

\begin{table}
\centering
\scalebox{0.7}{
\begin{tabular}{@{}l@{\hspace{8pt}}r@{\hspace{8pt}}r@{\hspace{6pt}}|l@{\hspace{10pt}}r@{\hspace{8pt}}r@{}}
\toprule
\textbf{Model} & \textbf{Start} & \textbf{End} & \textbf{Model} & \textbf{Start} & \textbf{End} \\ 
\midrule \midrule
\multicolumn{6}{c}{\textit{Model Size}} \\
\midrule
Qwen2.5-0.5B & 33.59 & 12.89 & Gemma3-0.27B & 19.92 & 37.89 \\
 Qwen2.5-1.5B & 32.42 & 13.67 & Gemma3-1B & 28.12 & 37.89 \\
 Qwen2.5-3B & 47.27 & 46.48 & Gemma3-4B & 55.47 & 35.94 \\
 Qwen2.5-7B & 66.41 & 67.19 & Gemma3-12B & 59.77 & 72.66 \\
 Qwen2.5-14B & 73.05 & 75.39 & Gemma3-27B & 85.55 & 87.89 \\
 Qwen2.5-32B & 56.25 & 74.22 & \cellcolor{gray!15}Ours (0.6B) & \cellcolor{gray!15}95.73 & \cellcolor{gray!15}93.02 \\
\midrule
\multicolumn{6}{c}{\textit{Model Types}} \\
\midrule
 Mistral-7B-Inst. & 79.69 & 81.64 & DeepSeekMath-7B & 66.41 & 69.14 \\
 Llama3.1-8B-Inst. & 73.83 & 74.61 & MetaMath-7B & 46.09 & 63.28 \\
 Qwen2.5-7B-Inst. & 53.12 & 81.64 & DeepSeek-R1-7B & 55.86 & 57.42 \\
 Gemma2-9B-Inst. & 89.06 & 86.72 & \cellcolor{gray!15}Ours (0.6B) & \cellcolor{gray!15}95.73 & \cellcolor{gray!15}93.02 \\
\bottomrule
\end{tabular}}
\vspace{-0.2cm}
\caption{\textbf{TIE accuracy of few-shot prompted LLMs vs. our
fine-tuned model.} Start/End denote accuracy on the predicted \texttt{start}/\texttt{end} (see Table~\ref{tbl:tie_task_format}). The exact prompts and zero-shot results are in \S\ref{supp:exp_tie}.}
\vspace{-0.05cm}
\label{tbl:tie_few_shot}
\end{table}

\subsection{TIE Model Analysis}
\label{subsec:exp_tie_task}

\paragraph{TIE Performance.}
To analyze the effectiveness of our trained TIE model (\S\ref{subsubsec:tie_model}), we compare
it against LLMs of varying sizes and types.
Table~\ref{tbl:tie_few_shot} shows that our 0.6B model demonstrates comparable performance to much larger models such as Gemma3-27B and instruction-tuned
models such as Gemma2-9B-Inst., confirming the effectiveness of
high-quality task-specific supervision from our constructed
dataset (\S\ref{subsubsec:tie_data}). This result also highlights \ours{}'s practicality under
memory or inference-time constraints, where deploying large
models is often infeasible. Further analysis (\S\ref{supp:exp_tie_tir}) confirms that gains on TIE
consistently translate into TIR improvements, with performance saturating
once TIE becomes sufficiently accurate.

\paragraph{Cost-Performance Tradeoff in TIR.}
While our TIE model itself is compact, integrating it as a
re-ranker (\S\ref{subsec:temporal_scoring}) incurs additional inference cost over plain retrieval. We thus analyze whether this
overhead is compensated by the resulting TIR performance gains.
Table~\ref{tbl:cost_performance} shows that \ours{} adds only
1.02--7.15\% latency on top of non-temporal retrievers, while
yielding substantial 22.97--24.44\%p NDCG@1 gains. We observe this favorable
tradeoff consistently across different retrievers.

\begin{table}[t!]
\centering
\scalebox{0.7}{
\begin{tabular}{@{}l|r@{\hspace{6pt}}r|r@{\hspace{6pt}}r|r@{\hspace{6pt}}r@{}}
\toprule
\textbf{Method} & \multicolumn{2}{c|}{\textbf{Latency (s)}} & \multicolumn{2}{c|}{\textbf{NDCG@1}} & \multicolumn{2}{c}{\textbf{NDCG@5}} \\
       & Value & $\Delta$ & Value & $\Delta$ & Value & $\Delta$ \\
\midrule \midrule
Contriever     & 34.97 & --              & 57.24 & --                & 70.94 & -- \\
\rowcolor{gray!15}
+ \ours{}      & 37.47 & \red{+7.15\%}   & 81.45 & \green{+24.21\%p} & 86.14 & \green{+15.20\%p} \\
\midrule
JinaAI         & 46.35 & --              & 61.65 & --                & 74.70 & -- \\
\rowcolor{gray!15}
+ \ours{}      & 46.83 & \red{+1.02\%}   & 86.09 & \green{+24.44\%p} & 89.48 & \green{+14.78\%p} \\
\midrule
NomicAI        & 67.53 & --              & 62.78 & --                & 76.98 & -- \\
\rowcolor{gray!15}
+ \ours{}      & 68.90 & \red{+2.02\%}   & 85.75 & \green{+22.97\%p} & 89.08 & \green{+12.10\%p} \\
\bottomrule
\end{tabular}}
\vspace{-0.2cm}
\caption{\textbf{Cost-performance tradeoff of \ours{}.} We
report inference latency (seconds) and retrieval performance (NDCG@$\{1,5\}$) when
\ours{} is applied on top of three different non-temporal
retrievers (Contriever, JinaAI, NomicAI). The $\Delta$ columns show relative changes.}
\label{tbl:cost_performance}
\end{table}

\begin{figure}[t!]
\vspace{-0.1cm}
\centering
\includegraphics[width=\columnwidth]{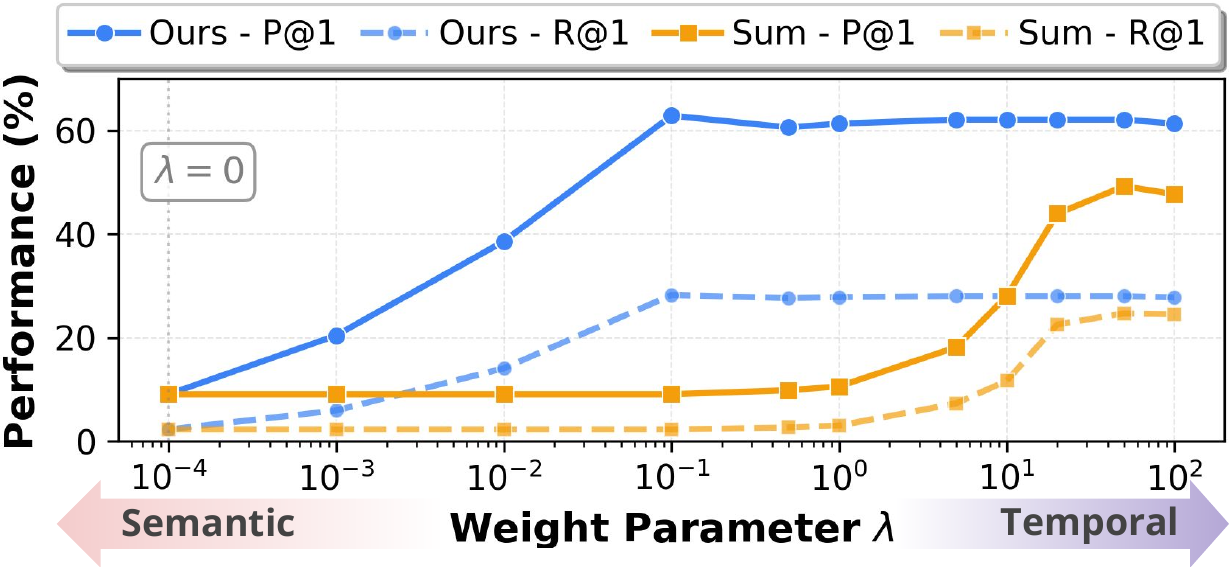}
\vspace{-0.6cm}
\caption{\textbf{TIR performance of PoE vs. weighted-sum aggregation across varying $\lambda$.} 
We report Precision@1 and Recall@1 of our PoE-based fusion (blue) and the weighted-sum baseline (orange), 
where lower/higher $\lambda$ emphasizes the semantic/temporal score.}
\label{fig:poe_ablation}
\end{figure}

\subsection{Ablation Study of PoE-based Fusion}

We conduct an ablation study to verify the effectiveness of our PoE-based fusion (\S\ref{subsec:poe_aggregation}). We compare retrieval performance against a standard weighted-sum aggregation as we sweep the weight parameter $\lambda$, which controls the strengths of the semantic and temporal scores (Eq~\ref{equ:poe_fusion}). Fig.~\ref{fig:poe_ablation} shows that our PoE fusion outperforms the weighted sum across nearly the entire range of $\lambda$, achieving up to 62.88\% P@1 compared to at most 49.24\% for the sum baseline. In particular, while the weighted sum collapses when the temporal signal is weak, PoE remains robust due to its multiplicative nature, validating it as a more reliable fusion mechanism for TIR.

\subsection{Integration in RAG Scenarios}
To further assess \ours{}'s downstream utility, we evaluate it within a 
temporal RAG pipeline on TempRAGEval, where the 
retrieved documents are fed to a reader model (Llama3.1-8B-Inst) to generate 
the final answer. We compare \ours{} against three categories of RAG baselines: 
\textbf{(1)} \textit{Standard RAG}, which retrieves documents once and directly 
feeds them to the reader; 
\textbf{(2)} \textit{Re-rankers}, which apply an additional neural re-ranker model after the initial retrieval; and 
\textbf{(3)} \textit{Self-RAG}~\citep{asai2024self}, an advanced RAG method that 
performs multiple retrieval steps guided by the reader model. 
As shown in Table~\ref{tbl:exp_rag}, \ours{} outperforms the standard RAG baseline and matches the much larger neural re-ranker (Qwen3-4B) while being nearly 
$19\times$ faster (154.64s vs. 2924.31s) due to its compact size. 
These results show that \ours{} offers a simple yet effective 
drop-in re-ranker that improves temporal RAG performance 
without sacrificing efficiency.

\begin{table}[t!]
\centering
\scalebox{0.7}{
\begin{tabular}{@{}l|l|r@{\hspace{12pt}}r@{\hspace{8pt}}r@{}}
\toprule
\textbf{Category} & \textbf{Method} & \textbf{F1} & \textbf{Acc} & \!\textbf{Latency(s)} \\ 
\midrule \midrule
Standard RAG & Contriever & 19.67 & 32.86 & 159.52  \\
Standard RAG & JinaAI & 22.67 & 38.96 & 150.48  \\
Standard RAG & TempRetriever & 16.94 & 33.80 & 171.17  \\
Standard RAG & TsContriever & 24.54 & 42.25 & \underline{141.74}  \\
Standard RAG & TSM &  22.46 & 43.19 & \textbf{139.11} \\
\midrule
Self-RAG & Contriever & 16.12 & 40.85 & 916.20  \\
Self-RAG & JinaAI & 18.00 & 43.19 & 952.59  \\
Re-rank \small{(Model-based)} & Qwen3-0.6B & 22.59 & 44.13 & 645.85  \\
Re-rank \small{(Model-based)} & Qwen3-4B & \textbf{26.88} & \textbf{49.76} & 2924.31 \\

\midrule
\rowcolor{gray!15}
Standard RAG & \ours{} & \underline{24.80} & \underline{45.07} & 154.64 \\
\bottomrule
\end{tabular}}
\vspace{-0.1cm}
\caption{\textbf{RAG performance on TempRAGEval}. We report F1 score and answer accuracy for RAG performance and latency for computational efficiency.}
\vspace{-0.1cm}
\label{tbl:exp_rag}
\end{table}

\begin{table}[t!]
\centering
\scalebox{0.7}{
\begin{tabular}{@{}l@{\hspace{10pt}}|p{5cm}@{\hspace{8pt}}|c@{\hspace{3pt}}c@{}}
\toprule
\textbf{Retriever} & \textbf{Top-1 Retrieved Document} & \textbf{Sem.} & \textbf{Temp.} \\ 
\midrule \midrule
\multicolumn{4}{@{}l}{\textit{$q_1$: Who was the winner of the Nobel Peace Prize in 1926?}} \\
\midrule
TsContriever & Gustav Stresemann. The Nobel Peace Prize 1926. Born\ldots                     & \cmark & \cmark \\
\ours{}      & Aristide Briand. The Nobel Peace Prize 1926. Born\ldots          & \cmark & \cmark \\
\midrule
\multicolumn{4}{@{}l}{\textit{$q_2$: Who won the Nobel Prize in Chemistry between 1924 and 1925?  }} \\
\midrule
TsContriever & Fritz Pregl. The Nobel Prize in Chemistry 1923. Born\ldots                     & \cmark & \xmark \\
\ours{}      & Richard Zsigmondy. The Nobel Prize in Chemistry 1925. Born\ldots          & \cmark & \cmark \\
\midrule
\multicolumn{4}{@{}l}{\textit{$q_3$: Who won the Nobel Prize in Chemistry one year before 1926?  }} \\
\midrule
TsContriever & Johannes Fibiger. The Nobel Prize in Medicine 1926. Born\ldots                     & \xmark & \xmark \\
\ours{}      & Richard Zsigmondy. The Nobel Prize in Chemistry 1925. Born\ldots          & \cmark & \cmark \\
\bottomrule
\end{tabular}}
\vspace{-0.2cm}
\caption{\textbf{Examples of top-1 retrieved documents on temporal queries.} 
We compare \ours{} against TsContriever on three queries of increasing complexity of temporal constraints. Sem./Temp denotes semantic/temporal relevance to the query. For a balanced view, we also examine when \ours{} fails in
\S\ref{supp:exp_failure_analysis}.}
\label{tbl:qualitative_comparison}
\end{table}

\subsection{Case Study}
\label{subsec:exp_case_study}
To better understand the source of \ours{}'s gains, we qualitatively compare 
its top-1 retrievals against TsContriever on three queries of increasing 
temporal complexity. As shown in Table~\ref{tbl:qualitative_comparison}, 
both retrievers succeed on a simple explicit expression ($q_1$), but 
TsContriever fails on range-based ($q_2$) and relative ($q_3$) temporal 
expressions. Notably, on the most challenging query ($q_3$), its temporal 
misinterpretation cascades into a semantic error: by overfocusing on the year 
``1926,'' it retrieves a document from an entirely different category. 
In contrast, \ours{} consistently retrieves documents satisfying both 
semantic and temporal relevance. For a balanced view, we further analyze when \ours{} fails in
\S\ref{supp:exp_failure_analysis}.

\subsection{More Analysis}
\label{subsec:exp_more_analysis}
We further analyze \ours{} along the following axes.

\paragraph{Extension to More Complex Temporal Queries.}
\ours{} targets queries whose temporal intent can be normalized into a time
interval from the query itself. However, real-world queries can be more
diverse, often expressing \textit{implicit} constraints that require
background knowledge to resolve~\citep{su2024temporal}. For example,
\textit{``During the 23rd Winter Olympics, who was the U.S. president?''}
demands knowing when the Olympics took place. To handle such queries, many
RAG strategies like query rewriting~\citep{qian2024timer4} have been
developed to convert implicit references into explicit conditions (e.g.,
\textit{the 23rd Winter Olympics} $\rightarrow$ \textit{February 2018}).
Although such implicit queries lie beyond \ours{}'s scope alone, our
experiments in \S\ref{supp:multi_event} show that combining \ours{} with
query rewriting yields strong performance on implicit queries (+12.89 Hit). While even
more challenging temporal queries can exist, these results
suggest that precisely resolving explicit constraints -- where \ours{}
contributes -- is a critical prerequisite toward these broader settings.

\paragraph{When Non-temporal Queries are Mixed.}
While \ours{} is designed for temporal queries, real-world corpora such as
news archives often mix temporal and non-temporal queries. To investigate
\ours{}'s generalization to this mixed setting, we evaluate it on
ChroniclingQA~\citep{piryani2024chroniclingamericaqa}, a benchmark built on
historical American newspapers, where the dynamic
nature of news makes queries inherently temporal while non-temporal queries
are also mixed in. Analysis in \S\ref{supp:non_temporal} shows that on a balanced mixture of
temporal and non-temporal queries, \ours{} improves over the non-temporal
retriever (NDCG@1: 47.66 $\rightarrow$ 50.86). Even on purely non-temporal
queries, \ours{} incurs only a marginal drop (NDCG@10: 72.62 vs.\ 70.27),
as the strength of the temporal score is tunable via a weight
hyperparameter. These results are
consistent with our main findings on Nobel Prize and TempRAGEval, supporting
\ours{}'s applicability to real-world corpora.

\paragraph{Robustness to Missing or Noisy Document Timestamps.}
While our main experiments assume document timestamps exist, such metadata is
often unavailable or unreliable in real-world corpora. To investigate
\ours{}'s robustness in this setting, we
replace gold timestamps with noisy intervals estimated from dates mentioned
in each document's text. Even under this noise, analysis in \S\ref{supp:noisy_doc_time} show that \ours{} achieves the best
performance among temporal re-rankers by large margins (e.g., P@1: 80.19
vs.\ at most 63.68 on the simple subset), and surpasses jointly-trained
temporal retrievers on the complex subset, suggesting practical
applicability to corpora with imperfect temporal metadata.

\paragraph{Statistical Significance of Improvements.}
Following IR practice, we perform paired per-query significance
tests between \ours{} and all baselines on an OOD setting (\S\ref{supp:stat_sig}). We observe \ours{}'s
improvements are statistically significant in 246 out of 270 comparisons
($p\!<\!0.05$), and every P@1 and NDCG@5 improvement is significant at
$p\!<\!0.001$.

\section{Conclusion}
We presented \ours{}, a lightweight temporal re-ranking framework that bridges a common trade-off in existing TIR paradigms: strong temporal constraint enforcement versus flexible understanding of diverse temporal expressions. To this end, we introduced Temporal Intent Extraction (TIE) as a learned task that maps diverse temporal expressions in queries into a unified \tempint{} representation. This formulation allows \ours{} to explicitly penalize temporal mismatch based on the identified interval, while remaining flexible to diverse and unseen expressions via model-based learning. We constructed a dedicated dataset to train a specialized TIE model and used a Product-of-Experts aggregation to better balance semantic and temporal relevance. Experiments demonstrated that \ours{} consistently outperforms prior TIR methods across diverse query types, with its 0.6B TIE model achieving strong performance at substantially lower inference cost than larger alternatives. \ours{} further integrates seamlessly with existing RAG pipelines, offering a practical and extensible module for integrating temporal awareness.

\section*{Limitations}
\vspace{-0.1cm}

We discuss several directions that remain open for future work. First, \ours{} currently operates at month-level granularity. While this is finer than the year-level granularity typically adopted in prior TIR, it may still fall short in domains requiring day-level retrieval (e.g., news or financial corpora). Second, we primarily investigated \ours{} in retrieval settings where temporal scoring is performed once per query, following the commonly used TIR evaluation setup. Extending \ours{} and other TIR methods to more sophisticated RAG pipelines that require iterative, reader-guided temporal scope refinement is a promising direction. Finally, while our evaluation focuses on document retrieval, generalizing \ours{} to structured data such as temporal knowledge graphs would provide a more complete picture of its applicability.

\section*{Acknowledgments}
This work was supported by the NYU-KAIST Partnership and by the IITP with a grant funded by the Ministry of Science and ICT (MSIT) of the Republic of Korea in connection with the Global AI Frontier Lab International Collaborative Research. (No. RS-2024-00469482 \& RS-2024-00509258). This work was also supported by the National Research Foundation of Korea (NRF) grant funded by the Korea government (MSIT) (No.\@ RS-2022-NR070121).

\bibliography{main_emnlp}

\clearpage
\appendix

\section{Appendix -- TIE Task}
\label{supp:tie_task}
Continuing from \S\ref{subsec:tie}, we provide more details on the TIE dataset construction and the TIE model training process.

\subsection{TIE Data Construction}
\label{supp:tie_data_process}

\paragraph{Parsing SQL Constraints from TDBench.}
TDBench~\citep{kim2025harnessing} pairs each natural 
language query with a semantically equivalent SQL query 
that encodes its temporal constraints over \texttt{Start} 
and \texttt{End} columns. We implement a rule-based Python 
parser that converts each SQL constraint into a temporal 
window $\langle t_{\text{start}}, t_{\text{end}} \rangle$, 
where either bound may be \texttt{None} when the constraint 
is one-sided. Table~\ref{tab:sql_parsing} shows 
representative cases:

\begin{itemize}[leftmargin=0.5cm]
\vspace{-0.25cm}
\setlength\itemsep{-1pt}
    \item \textit{Two-sided constraints} 
    (e.g., \texttt{Start between `YYYY-01-01' and 
    `YYYY-12-31' and End between ...}) yield both bounds.
    \item \textit{One-sided constraints} 
    (e.g., \texttt{Start between `2006-03-01' and 
    `2006-03-31'}) yield a single bound, with the other set 
    to \texttt{None}.
    \item \textit{Relative-date constraints} 
    (e.g., \texttt{date(End) = date(`2019-03-09', `-8 
    month')}) are resolved by evaluating the SQL date 
    arithmetic in Python.
\end{itemize}

\paragraph{Manual Annotation for Nobel Prize.}
The original Nobel Prize benchmark~\citep{wu2024time} only features basic temporal expressions (e.g., \textit{in}, \textit{between}), which do not cover the full range of Allen's interval relations as in TDBench. To address this, the authors manually wrote 150 additional queries with more complex constraints (e.g., \textit{before}, \textit{after}, \textit{N years prior to}). These complex queries were designed to preserve the original ground-truth temporal windows by reformulating the temporal expressions (e.g., transforming \textit{``in 2017''} into \textit{``3 years prior to 2020''}).

\begin{table*}[t!]
\centering
\small
\begin{tabular}{p{6.5cm}p{5cm}p{3cm}}
\toprule
\textbf{Natural Language Query} & \textbf{SQL Constraint} & \textbf{Temporal Window} \\
\midrule \midrule
During the period from 2006 to 2007, which basketball team was Kevin Durant a part of? & 
\texttt{Start between `2006-01-01' and `2006-12-31' and End between `2007-01-01' and `2007-12-31'} & 
start: 2006-01 \newline end: 2008-01 \\ 
\midrule
Who was the President of Portugal that started their term in March 2006? & 
\texttt{Start between `2006-03-01' and `2006-03-31'} & 
start: 2006-03 \newline end: None \\ 
\midrule
Who was the Prime Minister of Turkey whose term ended exactly 8 months before March 9, 2019? & 
\texttt{date(End) = date(`2019-03-09', `-8 month')} & 
start: None \newline end: 2018-07 \\
\bottomrule
\end{tabular}
\vspace{-0.2cm}
\caption{\textbf{Examples of temporal window extraction from 
TDBench SQL constraints}. We use a rule-based Python 
parser to convert each SQL constraint into a temporal 
window \tempint{}.}
\label{tab:sql_parsing}
\end{table*}

\begin{table*}[t!]
\centering
\small
\begin{tabular}{p{0.46\linewidth} @{\hspace{0.06\linewidth}} p{0.46\linewidth}}
\toprule
\textbf{Training Prompt} & \textbf{Inference Prompt} \\
\midrule \midrule
\verb|### Instruction | \newline
\verb|Extract the time span from the question.| \newline
\verb|Answer format is below two lines.| \newline
\verb|start: YYYY-MM or None.| \newline
\verb|end: YYYY-MM or None.| \newline
\newline
\verb|### Input| \newline
\verb|Question: <query>| \newline
\newline
\verb|### Response| \newline
\verb|Answer:| \newline
\verb|start: <start>| \newline
\verb|end: <end>| 
& 
\verb|### Instruction| \newline
\verb|Extract the time span from the question.| \newline
\verb|Answer format is below two lines.| \newline
\verb|start: YYYY-MM or None.| \newline
\verb|end: YYYY-MM or None.| \newline
\newline
\verb|### Input| \newline
\verb|Question: <query>| \newline
\newline
\verb|### Response| \newline
\verb|Answer:| \\
\bottomrule
\end{tabular}
\vspace{-0.2cm}
\caption{\textbf{Training and inference prompts used for the TIE task} (Table~\ref{tbl:tie_task_format}). \texttt{<query>}, \texttt{<start>}, and \texttt{<end>} denote placeholders for a query and its temporal window.}
\label{tbl:tie_prompts}
\end{table*}

\begin{table*}[t!]
\centering
\small
\begin{tabular}{l@{\hspace{20pt}}cccccc}
\toprule
\textbf{Dataset} & \textbf{\# Query} & \textbf{\# Simple} & \textbf{\# Complex} & \textbf{\# Documents} & \textbf{Avg. Docs/Query} & \textbf{Avg. Tokens/Doc} \\
\midrule \midrule
Nobel Prize  & 3,399 & 3,244 & 155 & 989   & 3.7 & 72.4 \\
TempRAGEval  & 884 & 212 & 672 & 9,695 & 1.0 & 165.2 \\
\bottomrule
\end{tabular}
\vspace{-0.2cm}
\caption{\textbf{Statistics of Evaluation Datasets.} For each dataset (Dataset), we report the total number of queries (\# Query), the number of queries in the simple (\# Simple) and complex (\# Complex) subsets, the number of unique documents (\# Documents), the average number of relevant documents per query (Avg. Docs/Query), and the average document length in tokens (Avg. Tokens/Doc) measured via the BERT tokenizer.}
\label{tab:dataset_stats}
\end{table*}

\subsection{TIE Model Training}
\label{supp:tie_training}

\paragraph{Training and Inference Prompts.}
Table~\ref{tbl:tie_prompts} shows the prompt templates used during training and inference. Both follow an Instruction-Query-Response structure, where the response contains the target time span required in the query (\texttt{start} and \texttt{end} formatted as \texttt{YYYY-MM}, or \texttt{None}). During inference, the model is prompted with the same template truncated at the \texttt{Answer:} marker, and the time span is parsed from the generated continuation.

\paragraph{Training Details.}
During supervised fine-tuning (SFT), we train the model using a standard causal language-modeling loss over the full sequence with the EOS token appended to the answer. For parameter-efficient tuning, we apply LoRA with rank $r{=}8$ and $\alpha{=}16$. We optimize the model for $4$ epochs using the AdamW optimizer with a learning rate of $2{\times}10^{-4}$ and a batch size of $16$. To support reproducibility, we will release both the training script and the trained model checkpoints.

\section{Appendix -- PoE Formulation}
\label{supp:poe_formulation}

Continuing from \S\ref{subsec:poe_aggregation}, we provide a detailed derivation of the Product-of-Experts (PoE) formulation in Eq.~\ref{equ:poe_fusion}. 

Since we operate in a retriever-agnostic setting without direct access to the base retriever's internal probabilistic framework, we follow energy-based models~\citep{lecun2006tutorial} to interpret the raw semantic score $S_{\text{sem}}(q, d)$ as an unnormalized log-probability:
\begin{equation}
S_{\text{sem}}(q, d) \approx\log P_{\text{sem}}(q, d) + C,
\end{equation}
where $P_{\text{sem}}(q, d)$ represents the relevance probability estimated by the semantic expert, and $C$ is a normalization constant. 

For the temporal scoring mechanism, our design ensures that $S_{\text{temp}}(q, d)$ is already bounded within $[0, 1]$ (see \S\ref{subsec:temporal_scoring}). We can therefore interpret this score directly as the relevance probability from the temporal expert, i.e., $P_{\text{temp}}(q, d) = S_{\text{temp}}(q, d)$. 

By substituting these probabilistic interpretations into our weighted sum function, the final ranking score expands as follows:
\begin{align}
&S_{\text{final}}(q, d) = S_{\text{sem}}(q, d) + \lambda \cdot \log S_{\text{temp}}(q, d) \nonumber \\
&\approx \left( \log P_{\text{sem}}(q, d) + C \right) + \lambda \cdot \log P_{\text{temp}}(q, d) \nonumber \\
&= \log P_{\text{sem}}(q, d) + \log P_{\text{temp}}^{\lambda}(q, d) + C \nonumber \\
&= \log \left( P_{\text{sem}}(q, d) \cdot P_{\text{temp}}^{\lambda}(q, d) \right) + C.
\end{align}

This log-space addition mathematically corresponds to a multiplicative fusion ($P_{\text{sem}} \cdot P_{\text{temp}}^{\lambda}$) in the probability space. Consequently, it enforces a strict joint constraint where \textit{both} experts must yield high probabilities. Unlike a standard linear weighted sum common in TIR -- where an exceptionally high semantic score can override a temporal mismatch -- our formulation heavily penalizes the final score as $\log S_{\text{temp}}(q, d)$ approaches negative infinity when the temporal constraint is violated ($S_{\text{temp}} \to 0$).

\section{Appendix -- Experimental Setup}
\label{supp:exp_setup}
Continuing from \S\ref{subsec:exp_setup}, we provide more details on experimental setups. All experiments are performed on NVIDIA Quadro RTX 8000 GPUs.

\subsection{Statistics of Evaluation Datasets}
\label{supp:exp_setup_data_stat}
Continuing from \S\ref{subsec:exp_setup}, we report detailed
statistics of the datasets used to evaluate \ours{}. Following
the TempRAGEval criterion, queries are classified as \emph{simple}
if their temporal expression is explicit with a direct year
mention that also appears in the document, and \emph{complex}
otherwise. Table~\ref{tab:dataset_stats} summarizes the number of
queries in each subset, along with the number of unique
documents, average number of relevant documents per query, and
average document length in tokens measured via the BERT
tokenizer\footnote{https://huggingface.co/google-bert/bert-base-uncased}.

\subsection{Baseline Details}
\label{supp:exp_setup_baselines}
Continuing from \S\ref{subsec:exp_setup}, we provide the full
list of baselines used in our experiments, following the five
categories introduced in the main text.

\paragraph{Non-temporal retrievers.}
We use BM25~\citep{robertson2009probabilistic},
Contriever~\citep{izacard2021unsupervised},
JinaAI~\citep{gunther2023jina}, and NomicAI~\citep{nussbaum2024nomic}. These
baselines measure semantic relevance without explicit temporal
modeling.

\paragraph{Temporal retrievers.}
We use TsContriever~\citep{wu2024time},
TempRetriever~\citep{abdallah2025tempretriever}, and TSM~\citep{han2025temporal}.
These models jointly learn semantic and temporal relevance
through fine-tuning on temporally supervised data.

\paragraph{Temporal re-rankers.}
We use TempRALM~\citep{gade2025s}, MRAG~\citep{siyue2024mrag}, and
TimeR4~\citep{qian2024timer4}. These methods apply rule-based or
learned temporal scoring on top of a fixed retriever. While MRAG
and TimeR4 are originally proposed as end-to-end RAG frameworks,
we isolate their score-based re-ranking components  for fair comparison,
following the official implementations released by the authors.

\paragraph{General neural re-rankers.}
We use Qwen3-Embedding-0.6B and Qwen3-Embedding-4B~\citep{yang2025qwen3}, which represent the
state-of-the-art in general-purpose neural re-ranking, but lack
any temporal specialization. Comparing against them helps
quantify how much temporal modeling adds beyond strong semantic
matching.

\paragraph{General-purpose LLMs.}
For the TIE task, we additionally compare against LLMs spanning a wide range of sizes (0.5B--32B) and training
paradigms: \textit{general instruction-tuned models}
(Mistral-7B-Inst.~\citep{jiang2023mistral},
Llama3.1-Inst.~\citep{grattafiori2024llama},
Qwen2.5-Inst.~\citep{qwen2.5},
Gemma2-Inst.~\citep{team2024gemma}); \textit{reasoning-focused
models} (DeepSeekMath-7B~\citep{shao2024deepseekmath},
MetaMath-7B~\citep{yu2023metamath},
DeepSeek-R1~\citep{guo2025deepseek}); and \textit{base models}
(Qwen2.5~\citep{qwen2.5}, Gemma3~\citep{team2025gemma}). All
models are prompted with the same instruction template (see Table~\ref{tbl:tie_prompts}), and performance is measured by
exact-match accuracy on both the start and end dates of the
predicted temporal window.

\subsection{Metric Definitions}
\label{supp:exp_setup_metrics}

Continuing from \S\ref{subsec:exp_setup}, we provide formal
definitions of Precision@$k$, Recall@$k$, and NDCG@$k$, Hit, F1 score, and answer accuracy, which are the
standard retrieval and RAG evaluation metrics~\citep{robertson2009probabilistic} used in our
experiments. All metrics are computed on
a per-query basis and then averaged across all queries to obtain
the final reported values. Below, $d_i$ denotes the
$i$-th retrieved document, and $\mathcal{D}_{\text{rel}}(q)$ denotes the set of
all relevant documents for query $q$.

\begin{itemize}[leftmargin=10pt]
\vspace{-0.1cm}
    \item \textit{Precision@$k$:} The proportion of
    relevant documents among the top-$k$ retrieved documents for
    each query.
    \begin{equation}
        \text{Precision@}k = \frac{1}{k} \sum_{i=1}^{k} \mathds{1}[d_i \text{ is relevant}].
    \end{equation}
    \item \textit{Recall@$k$:} The proportion of all
    relevant documents that are retrieved in the top-$k$ results
    for each query.
    \begin{equation}
        \text{Recall@}k = \frac{1}{|\mathcal{D}_{\text{rel}}(q)|} \sum_{i=1}^{k} \mathds{1}[d_i \text{ is relevant}].
    \end{equation}
    \item \textit{Normalized Discounted Cumulative Gain@$k$
    (NDCG@$k$):} A ranking quality metric that accounts for the
    positions of relevant documents within the top-$k$ results,
    giving higher weights to relevant documents appearing earlier
    in the ranking.
    \begin{align}
        \text{DCG@}k &= \sum_{i=1}^{k} \frac{2^{\mathds{1}[d_i
    \text{ is relevant}]} - 1}{\log_2(i+1)}, \\
        \text{NDCG@}k &= \frac{\text{DCG@}k}{\text{IDCG@}k}.
    \end{align}
    Here, $\text{IDCG@}k$ is the maximum
    possible DCG for the ideal ordering.

    \item \textit{Hit:} Whether the ground-truth answer appears
    as a substring of the LLM-generated answer.
    \begin{align}
        \text{Hit@}k = \mathds{1}\left[\text{Ground Truth} \subseteq \text{Predicted}\right],
    \end{align}.
    
    \item \textit{F1 score:} The token-level overlap between the LLM-generated answer and the ground-truth answer~\citep{rajpurkar2016squad}, computed from the precision and recall of predicted tokens.
    \begin{align}
        \text{Precision} &= \frac{|\text{Predicted} \cap \text{Ground Truth}|}{|\text{Predicted}|}, \\
        \text{Recall} &= \frac{|\text{Predicted} \cap \text{Ground Truth}|}{|\text{Ground Truth}|}, \\
        \text{F1} &= \frac{2 \cdot \text{Precision} \cdot \text{Recall}}{\text{Precision} + \text{Recall}}.
    \end{align}
    
    \item \textit{Answer Accuracy:} Whether the correct answer appears in the generated output. We adopt answer accuracy instead of exact match, as exact match does not reliably handle surface-form variations in final answers (e.g., \textit{Great Northern Railway} vs.\ \textit{GNR}).
    
\end{itemize}

\section{Appendix -- Experiments}
\label{supp:exp}
Continuing from \S\ref{sec:experiments}, we provide more experimental results and analysis.

\subsection{TIR Performance Evaluation}
Continuing from \S\ref{subsec:exp_tir_performance}, we provide additional NDCG metrics along with Precision, Recall metrics shown in Table~\ref{tbl:nobel_timeqa_retrieval}. Table~\ref{tbl:ndcg_retrieval} shows that the performance trends are consistent with Precision and Recall metrics; \ours{} achieves the best NDCG across all settings, with particularly large gains on the more
challenging Nobel Prize benchmark.

\subsection{TIE Performance Evaluation}
\label{supp:exp_tie}
Continuing from \S\ref{subsec:exp_tie_task}, we provide
additional results and details on the TIE model evaluation.

\paragraph{Zero-Shot Results.}
Table~\ref{tbl:tie_zero_shot} reports zero-shot TIE accuracy across the same LLMs evaluated under few-shot prompting in Table~\ref{tbl:tie_few_shot}. As expected, zero-shot results are generally lower than in few-shot, indicating that in-context demonstrations help calibrate the expected output format. However, the overall ranking among baselines remains largely consistent, suggesting that prompting strategies amplify but do not fundamentally alter each model's capability for the TIE task. In both settings, our fine-tuned 0.6B model achieves the highest accuracy, with an even larger margin under zero-shot, further supporting that task-specific supervision is more effective than prompting strategies alone.

\begin{table*}[]
\centering
\scalebox{0.7}{
\begin{tabular}{@{}l|r@{\hspace{8pt}}r@{\hspace{8pt}}r@{\hspace{8pt}}r| r@{\hspace{8pt}}r@{\hspace{8pt}}r@{\hspace{8pt}}r| r@{\hspace{8pt}}r@{\hspace{8pt}}r@{\hspace{8pt}}r| r@{\hspace{8pt}}r@{\hspace{8pt}}r@{\hspace{8pt}}r@{\hspace{4pt}}|}
\toprule
 & \multicolumn{8}{c|}{\textbf{Nobel Prize}} & \multicolumn{8}{c|}{\textbf{TempRAGEval}} \\
 & \multicolumn{8}{c|}{Simple\hspace{100pt}Complex}
 & \multicolumn{8}{c|}{Simple\hspace{100pt}Complex} \\
\cmidrule(lr){2-5}\cmidrule(lr){6-9}
\cmidrule(lr){10-13}\cmidrule(lr){14-17}
\textbf{Method}
& @1 & @5 & @10 & @50
& @1 & @5 & @10 & @50
& @1 & @5 & @10 & @50
& @1 & @5 & @10 & @50 \\
\midrule \midrule
\multicolumn{17}{c}{\textit{Non-temporal and Temporal Retrievers}} \\
\midrule
BM25
& 15.60 & 21.69 & 29.08 & 39.59
& 2.88 & 3.13 & 3.94 & 7.30
& 61.79 & 73.25 & 74.36 & 76.16
& 45.54 & 57.53 & 60.08 & 62.10 \\
Contriever
& 17.69 & 17.33 & 21.51 & 35.27
& 10.32 & 12.60 & 17.72 & 29.53
& 61.32 & 75.07 & 76.93 & 77.92
& 55.95 & 69.63 & 71.93 & 73.54 \\
JinaAI
& 38.47 & 36.34 & 41.40 & 53.35
& 15.87 & 20.06 & 24.40 & 35.56
& 65.57 & 78.09 & 79.48 & 80.91
& 60.42 & 73.64 & 75.84 & 76.83 \\
\midrule
TempRetriever
& 18.28 & 15.77 & 19.30 & 32.13
& 7.69 & 8.19 & 10.61 & 23.07
& 54.72 & 65.17 & 66.53 & 69.22
& 42.86 & 53.58 & 55.68 & 58.93 \\
TsContriever
& 58.63 & 54.77 & 60.09 & 68.91
& 13.22 & 21.95 & 27.71 & 40.79
& 77.36 & 87.08 & 87.89 & 88.21
& 46.43 & 64.51 & 66.68 & 68.65 \\
TSM
& 38.13 & 36.85 & 43.34 & 56.71
& 18.75 & 23.89 & 29.68 & 43.44
& 73.11 & 85.13 & 85.62 & 86.51
& 57.89 & 73.11 & 75.23 & 76.79 \\
\midrule
\multicolumn{17}{c}{\textit{Temporal Re-rankers}} \\
\midrule
TempRALM \small(+ Contriever)
& 0.43 & 0.53 & 0.80 & 10.47
& 25.24 & 23.38 & 25.50 & 28.39
& 5.66 & 8.45 & 8.78 & 8.78
& 26.04 & 35.41 & 35.91 & 36.12 \\
MRAG \small(+ Contriever)
& 51.82 & 46.10 & 53.66 & 59.81
& 14.66 & 20.47 & 25.23 & 35.11
& 62.26 & 72.81 & 74.35 & 76.42
& 47.02 & 58.56 & 60.81 & 64.52 \\
TimeR4 \small(+ Contriever)
& 17.69 & 17.33 & 21.51 & 35.27
& 17.31 & 19.91 & 24.38 & 36.60
& 59.91 & 72.79 & 74.79 & 75.58
& 48.96 & 63.60 & 65.91 & 67.49 \\
\midrule
TempRALM \small(+ JinaAI)
& 0.71 & 0.78 & 1.05 & 9.06
& \underline{32.45} & 32.26 & 33.95 & 36.99
& 8.02 & 9.51 & 9.81 & 9.81
& 34.08 & 39.95 & 40.36 & 40.47 \\
MRAG \small(+ JinaAI)
& \underline{66.89} & 56.06 & \underline{65.56} & \underline{71.02}
& 18.03 & 23.78 & 29.38 & 39.07
& 63.68 & 71.78 & 73.18 & 75.13
& 32.89 & 41.80 & 43.50 & 50.16 \\
TimeR4 \small(+ JinaAI)
& 38.47 & 36.34 & 41.40 & 53.35
& 21.39 & 27.33 & 32.82 & 42.73
& 65.57 & 76.37 & 77.76 & 79.19
& 57.74 & 70.08 & 72.04 & 72.78 \\
\midrule
\rowcolor{gray!15}
\ours{} \small(+ Contriever)
& 60.54 & \underline{60.91} & 65.46 & 68.34
& 31.25 & \underline{41.93} & \underline{46.40} & \underline{51.81}
& \underline{88.21} & \underline{92.76} & \underline{93.03} & \underline{93.03}
& \underline{79.32} & \underline{88.39} & \underline{90.18} & \textbf{91.07} \\
\rowcolor{gray!15}
\ours{} \small(+ JinaAI)
& \textbf{78.76} & \textbf{77.84} & \textbf{80.07} & \textbf{81.96}
& \textbf{46.88} & \textbf{56.46} & \textbf{59.28} & \textbf{62.53}
& \textbf{93.87} & \textbf{97.64} & \textbf{98.11} & \textbf{98.11}
& \textbf{84.08} & \textbf{90.62} & \textbf{91.07} & \textbf{91.07} \\
\bottomrule
\end{tabular}}
\vspace{-0.2cm}
\caption{\textbf{Overall retrieval performance on Nobel Prize and
TempRAGEval (NDCG).} We report NDCG@$\{1,5,10\}$ on the simple
and complex subsets of each dataset, complementing the
Precision and Recall results in
Table~\ref{tbl:nobel_timeqa_retrieval}. We compare
non-temporal retrievers (BM25, Contriever, JinaAI), temporal
retrievers (TempRetriever, TsContriever, TSM), and temporal
re-rankers (TempRALM, MRAG, TimeR4), where re-rankers and
\ours{} re-rank the top-$k$ documents retrieved by Contriever or
JinaAI. \textbf{Bold} and \underline{underline}
denote the best and second-best results in each column. }
\label{tbl:ndcg_retrieval}
\end{table*}

\begin{table}[t]
\centering
\scalebox{0.7}{
\begin{tabular}{@{}l@{\hspace{10pt}}r@{\hspace{8pt}}r@{\hspace{6pt}}|l@{\hspace{12pt}}r@{\hspace{8pt}}r@{}}
\toprule
\textbf{Model} & \textbf{Start} & \textbf{End} & \textbf{Model} & \textbf{Start} & \textbf{End} \\ 
\midrule \midrule
\multicolumn{6}{c}{\textit{Model Size}} \\
\midrule
Qwen2.5-0.5B & 33.59 & 16.41 & Gemma3-0.27B & 34.77 & 14.06 \\
 Qwen2.5-1.5B & 34.77 & 14.06 & Gemma3-1B & 34.77 & 14.06 \\
 Qwen2.5-3B & 41.80 & 38.67 & Gemma3-4B & 51.95 & 40.23 \\
 Qwen2.5-7B & 57.03 & 64.45 & Gemma3-12B & 53.91 & 50.39 \\
 Qwen2.5-14B & 64.84 & 51.56 & Gemma3-27B & 73.83 & 85.16 \\
 Qwen2.5-32B & 69.53 & 71.48 & \cellcolor{gray!15}Ours (0.6B) & \cellcolor{gray!15}95.73 & \cellcolor{gray!15}93.02 \\
\midrule
\multicolumn{6}{c}{\textit{Model Types}} \\
\midrule
 Mistral-7B-Inst. & 46.88 & 71.48 & DeepSeekMath-7B & 51.17 & 44.14 \\
 Llama3.1-8B-Inst. & 60.94 & 56.25 & MetaMath-7B & 30.47 & 17.58 \\
 Qwen2.5-7B-Inst. & 65.62 & 65.23 & DeepSeek-R1-7B & 45.31 & 53.12 \\
 Gemma2-9B-Inst. & 83.59 & 66.80 & \cellcolor{gray!15}Ours (0.6B) & \cellcolor{gray!15}95.73 & \cellcolor{gray!15}93.02 \\
\bottomrule
\end{tabular}}
\vspace{-0.2cm}
\caption{\textbf{TIE accuracy of zero-shot prompted LLMs vs. our
fine-tuned model.} Start/End denote accuracy on the predicted \texttt{start}/\texttt{end} (see Table~\ref{tbl:tie_task_format}). Few-shot results are in Table~\ref{tbl:tie_few_shot}.}
\label{tbl:tie_zero_shot}
\end{table}

\paragraph{Prompts.}
Table~\ref{tbl:tie_fewshot_prompt} shows the exact few-shot prompt used for
the TIE task, which provides four in-context demonstrations covering
representative temporal expressions (relative offsets, open-ended
ranges, explicit start/end dates, and bounded periods) before the
test query. Zero-shot prompt is shown in
Table~\ref{tbl:tie_prompts}.

\begin{table*}[]
\centering
\small
\scalebox{0.9}{
\begin{tabular}{p{0.95\linewidth}}
\toprule
\textbf{Few-shot Inference Prompt} \\
\midrule \midrule
\texttt{\#\#\# Instruction} \newline
\texttt{Extract the time span from the question.} \newline
\texttt{Answer format is below two lines.} \newline
\texttt{start: YYYY-MM or None.} \newline
\texttt{end: YYYY-MM or None.} \newline
\newline
\texttt{\#\#\# Input} \newline
\texttt{Question: Which economist was awarded the Nobel Prize four years prior to 2016?} \newline
\newline
\texttt{\#\#\# Response} \newline
\texttt{Answer:} \newline
\texttt{start: 2012-01} \newline
\texttt{end: 2012-12} \newline
\newline
\texttt{\#\#\# Instruction} \newline
\texttt{Extract the time span from the question.} \newline
\texttt{Answer format is below two lines.} \newline
\texttt{start: YYYY-MM or None.} \newline
\texttt{end: YYYY-MM or None.} \newline
\newline
\texttt{\#\#\# Input} \newline
\texttt{Question: Who won the Nobel Prize in Physics later than 2017?} \newline
\newline
\texttt{\#\#\# Response} \newline
\texttt{Answer:} \newline
\texttt{start: 2017-12} \newline
\texttt{end: None} \newline
\newline
\texttt{\#\#\# Instruction} \newline
\texttt{Extract the time span from the question.} \newline
\texttt{Answer format is below two lines.} \newline
\texttt{start: YYYY-MM or None.} \newline
\texttt{end: YYYY-MM or None.} \newline
\newline
\texttt{\#\#\# Input} \newline
\texttt{Question: Who served as the Prime Minister of the Netherlands starting after July 31, 2010,} \newline
\texttt{and ending their term in July 2024?} \newline
\newline
\texttt{\#\#\# Response} \newline
\texttt{Answer:} \newline
\texttt{start: 2010-08} \newline
\texttt{end: 2024-07} \newline
\newline
\texttt{\#\#\# Instruction} \newline
\texttt{Extract the time span from the question.} \newline
\texttt{Answer format is below two lines.} \newline
\texttt{start: YYYY-MM or None.} \newline
\texttt{end: YYYY-MM or None.} \newline
\newline
\texttt{\#\#\# Input} \newline
\texttt{Question: Which individuals were Prime Ministers of the Czech Republic during the period} \newline
\texttt{from 1993 to 1998?} \newline
\newline
\texttt{\#\#\# Response} \newline
\texttt{Answer:} \newline
\texttt{start: 1993-01} \newline
\texttt{end: 1998-12} \newline
\newline
\texttt{\#\#\# Instruction} \newline
\texttt{Extract the time span from the question.} \newline
\texttt{Answer format is below two lines.} \newline
\texttt{start: YYYY-MM or None.} \newline
\texttt{end: YYYY-MM or None.} \newline
\newline
\texttt{\#\#\# Input} \newline
\texttt{Question: <query>} \newline
\newline
\texttt{\#\#\# Response} \newline
\texttt{Answer:} \\
\bottomrule
\end{tabular}}
\vspace{-0.2cm}
\caption{\textbf{Few-shot prompt provided to baseline LLMs for the TIE task.}
The prompt provides four demonstrations covering different temporal
expressions (relative offset, open-ended range, explicit start/end dates,
and bounded periods), followed by the test question.
\texttt{<query>} denotes the placeholder for the test question.}
\label{tbl:tie_fewshot_prompt}
\end{table*}

\subsection{Ablation Study on the TIE Model}
\label{supp:exp_tie_model_ablation}

We also investigate how the choice of backbone model for TIE (\S\ref{subsubsec:tie_model}) affects retrieval performance by varying its size and type. As a result, Table~\ref{tbl:tie_backbone_ablation} shows that our default 0.6B model offers the best trade-off between cost and performance: the accuracy gap between 0.6B and 8B backbones is marginal (within ~3\%p on both Start and End), suggesting that our high-quality TIE dataset enables strong performance even with compact backbones, making large models unnecessary for this TIE task.

\begin{table}
\centering
\scalebox{0.75}{
\begin{tabular}{@{}l@{\hspace{8pt}}r@{\hspace{8pt}}r@{\hspace{6pt}}|l@{\hspace{10pt}}r@{\hspace{8pt}}r@{}}
\toprule
\textbf{Model} & \textbf{Start} & \textbf{End} & \textbf{Model} & \textbf{Start} & \textbf{End} \\
\midrule \midrule
\multicolumn{3}{c|}{\textit{Qwen3 Family}} & \multicolumn{3}{c}{\textit{Llama-3.2 Family}} \\
\midrule
Qwen3-1.7B & 94.23 & 92.31  & Llama3.2-1B & 93.85 & 91.92  \\
Qwen3-4B & 97.31 & 95.77 & Llama3.2-3B & 96.92 & 95.00 \\
Qwen3-8B & 97.69 & 94.23 & \cellcolor{gray!15}Ours (Qwen3-0.6B) & \cellcolor{gray!15}94.23 & \cellcolor{gray!15}92.69 \\
\bottomrule
\end{tabular}}
\vspace{-0.2cm}
\caption{\textbf{TIE accuracy across backbone model sizes and families.} All models are fine-tuned on our constructed TIE dataset with the same recipe (\S\ref{subsubsec:tie_model}). Start/End denote accuracy on the predicted \texttt{start}/\texttt{end} (see Table~\ref{tbl:tie_task_format}).}
\label{tbl:tie_backbone_ablation}
\end{table}

\subsection{TIE--TIR Performance Relationship}
\label{supp:exp_tie_tir}
To quantify the relationship between TIE accuracy and downstream TIR
performance, we vary TIE accuracy by evaluating checkpoints at different
training steps and measure the corresponding TIR performance on
TempRAGEval. As shown in Table~\ref{tbl:tie_tir}, TIR performance
consistently improves as TIE accuracy increases, and saturates once TIE
becomes sufficiently accurate ($\sim$88\%). This saturation likely stems from the distribution difference between the TIE
and TIR test sets: beyond a certain accuracy, the remaining TIE errors fall
outside the query distribution of TempRAGEval, and thus no longer affect
downstream retrieval.

\subsection{Failure Analysis of \ours{}}
\label{supp:exp_failure_analysis}

Complementing the case study in \S\ref{subsec:exp_case_study}, we examine
when \ours{} fails. Due to its two-stage design, temporal intent extraction (TIE)
followed by re-ranking (TIR), we identify two distinct failure modes.

\paragraph{Case 1: TIE fails, but TIR partially recovers.}
\begin{itemize}[leftmargin=0.5cm]
\vspace{-0.25cm}
\setlength\itemsep{-1pt}
\item Query: \textit{Which economist was awarded the Nobel Prize
four years prior to 2016?}
\item Gold interval: $\langle$2012-01, 2012-12$\rangle$ 
\item Extracted interval: $\langle$2011-12, 2012-01$\rangle$ (\red{\xmark})
\item Result: P@1 $=$ 0, \; P@3 $=$ 0.33
\end{itemize}
When TIE slightly misinterprets the start year (2011 vs.\ 2012), \ours{}
fails to retrieve the gold document at top-1 (P@1 = 0), yet succeeding within top-3 (P@3 = 0.33). This shows that
the Gaussian-smoothed scoring (Eq.~\ref{eq:time_score}) degrades gracefully
under small extraction errors rather than collapsing entirely.

\paragraph{Case 2: TIE succeeds, but TIR fails.}
\begin{itemize}[leftmargin=0.5cm]
\vspace{-0.25cm}
\setlength\itemsep{-1pt}
\item Query: \textit{When was the last time Mount Etna exploded as
of 2020?}
\item Gold interval: $\langle$\texttt{None}, 2020-12$\rangle$
\item Extracted interval: $\langle$\texttt{None}, 2020-12$\rangle$ ($\green{\checkmark}$)
\item Result: with the same $S_{\text{temp}} = 0.1017$, retrieval
succeeds or fails depending on the base retriever's semantic score:
\begin{itemize}[leftmargin=0.4cm]
\setlength\itemsep{-1pt}
\item Contriever: $S_{\text{sem}} = 0.3836$, retrieval fails
\item JinaAI: $S_{\text{sem}} = 0.7625$, retrieval succeeds
\end{itemize}
\end{itemize}
Even with a correct temporal interval, \ours{} cannot recover when the base
retriever assigns a poor semantic score, since PoE requires both signals to
be high. This reflects the trade-off of our modular design, where \ours{}
inherits the base retriever's semantic quality. We believe this trade-off is
favorable, as \ours{} directly benefits from emerging, stronger retrievers in the future.

\begin{table}
\centering
\small
\scalebox{0.95}{
\begin{tabular}{lccccc}
\toprule
\textbf{Training step} & \textbf{0} & \textbf{50} & \textbf{100} & \textbf{150} & \textbf{200} \\
\midrule \midrule
TIE accuracy & 59.58 & 77.50 & 87.92 & 88.33 & 92.91 \\
TIR accuracy & 70.93 & 71.95 & 74.89 & 74.10 & 74.66 \\
\bottomrule
\end{tabular}}
\vspace{-0.2cm}
\caption{\textbf{TIE--TIR relationship across training steps.} TIE accuracy
is measured on the TIE test set (Table~\ref{tbl:tie_stats}), and TIR accuracy (NDCG@1) on TempRAGEval (\S\ref{sec:experiments}).}
\label{tbl:tie_tir}
\end{table}

\subsection{Extension to More Complex Multi-Event Temporal Queries}
\label{supp:multi_event}
Continuing from \S\ref{subsec:exp_more_analysis}, we evaluate \ours{}'s extensibility on more complex real-world temporal queries. A notable case is \textit{multi-event} queries (e.g., \textit{``During the 23rd Winter Olympics, who was the U.S. president?''}), where resolving the query's temporal scope itself requires knowledge of an auxiliary event (i.e., \textit{the 23rd Winter Olympics}). Such queries fall outside the scope of the TIE task as currently formulated (\S\ref{subsec:tie}), since the temporal expression cannot be directly mapped to a unified interval without external knowledge about the referenced event.

We show, however, that \ours{} can be extended to such queries by integrating it with \textit{query rewriting}~\citep{qian2024timer4}, which leverages the reader model's knowledge to rewrite implicit event references into explicit dates (e.g., \textit{the 23rd Winter Olympics} $\rightarrow$ \textit{February 2018}). Once rewritten, the query becomes amenable to the TIE module, and \ours{} can be applied as in the standard setting. To evaluate this, we follow the Retrieve-Rewrite-Retrieve-Re-rank pipeline of TimeR4~\citep{qian2024timer4} and replace its final Retrieve-Re-rank stages with \ours{}, evaluating on the \textit{multiple} subset of the MultiTQ dataset~\citep{chen2023multi}, which contains 9,000 multi-event queries paired with 461,329 corpus triplets.

Table~\ref{tbl:rewriting_integration} reports the results. Without query rewriting, applying \ours{} directly to multi-event queries yields limited performance -- likely because, when semantic relevance and temporal context are tightly intertwined and implicitly expressed, explicitly enforcing decoupled temporal scoring may harm retrieval rather than help. With query rewriting, however, this limitation is effectively bypassed: \ours{} achieves the strongest performance on both Hit and F1 score (51.33/25.63), demonstrating \ours{}'s extensibility to more complex temporal queries when paired with appropriate RAG strategies.

\begin{table*}[t!]
\centering
\scalebox{0.8}{
\begin{tabular}{l@{\hspace{12pt}}r@{\hspace{12pt}}r@{\hspace{12pt}}r@{\hspace{12pt}}r@{\hspace{12pt}}r}
\toprule
\textbf{Method} & \textbf{NDCG@1} & \textbf{NDCG@3} & \textbf{NDCG@5} & \textbf{NDCG@10} & \textbf{NDCG@20} \\
\midrule \midrule
\multicolumn{6}{c}{\textit{Temporal:Non-temporal = 50:50}} \\
\midrule
Vanilla Retriever & 47.66 & 56.74 & 59.07 & 61.25 & 62.82 \\
+ \ours{} ($\lambda$=0.01, default) & \underline{50.47} & \textbf{59.29} & \underline{61.75} & \underline{63.70} & \underline{64.86} \\
\rowcolor{gray!15}
+ \ours{} ($\lambda$=0.002, optimized) & \textbf{50.86} & \underline{58.93} & \textbf{61.74} & \textbf{63.94} & \textbf{65.18} \\
\midrule
\multicolumn{6}{c}{\textit{Temporal:Non-temporal = 0:100}} \\
\midrule
Vanilla Retriever & \textbf{61.47} & \textbf{68.74} & \textbf{70.92} & \textbf{72.62} & \textbf{73.70} \\
+ \ours{} ($\lambda$=0.01, default) & 60.06 & 66.71 & 68.51 & 70.21 & 71.21 \\
\rowcolor{gray!15}
+ \ours{} ($\lambda$=0.03, optimized) & \underline{60.22} & \underline{66.77} & \underline{68.57} & \underline{70.27} & \underline{71.27} \\
\bottomrule
\end{tabular}}
\vspace{-0.2cm}
\caption{\textbf{Retrieval performance under different temporal:non-temporal 
query mixtures on ChroniclingQA.} We report NDCG metrics with different $\lambda$ values, which controls the strength of the temporal score during retrieval (Eq.~\ref{equ:poe_fusion}). We use Contriever as the underlying retriever. 
\textbf{Bold} and \underline{underline} denote the best and second-best 
results.}
\label{tbl:non_temporal}
\end{table*}

\begin{table}[t]
\centering
\scalebox{0.77}{
\begin{tabular}{@{}l|r@{\hspace{12pt}}r}
\toprule
\textbf{Pipeline} & \textbf{Hit} & \textbf{F1} \\ 
\midrule \midrule
\texttt{Retrieve} $\rightarrow$ \texttt{Re-rank}                                            & 48.11 & 21.64 \\
\texttt{Retrieve} $\rightarrow$ \ours{}                                            & 38.44 & 19.55 \\
\midrule
\rowcolor{gray!15}
\texttt{Retrieve} $\rightarrow$ \texttt{Rewrite} $\rightarrow$ \texttt{Retrieve} $\rightarrow$ \ours{}                     & \textbf{51.33} & \textbf{25.63} \\
\bottomrule
\end{tabular}}
\vspace{-0.2cm}
\caption{\textbf{Synergy with a query rewriting RAG strategy.} 
We adopt the Retrieve-Rewrite-Retrieve stages of TimeR4~\citep{qian2024timer4} and replace its final Re-rank stage with \ours{}, with the first two rows as baselines without query rewriting. We report RAG performance (Hit and F1) on the \textit{multiple} subset of MultiTQ, 
which contains multi-event temporal queries.}
\label{tbl:rewriting_integration}
\end{table}

\subsection{When Non-temporal Queries are Mixed}
\label{supp:non_temporal}

Continuing from \S\ref{subsec:exp_more_analysis}, we examine whether
\ours{} remains robust when temporal and non-temporal queries are
mixed. We evaluate TIR performance on
ChroniclingQA~\citep{piryani2024chroniclingamericaqa}, a benchmark
containing both temporal and non-temporal queries over the same news corpus
(1.2K documents). We consider two query mixtures: a balanced 50:50 mix of
temporal and non-temporal queries (1,282 queries total), and a 0:100 mix
consisting only of non-temporal queries (641 queries).

As shown in Table~\ref{tbl:non_temporal}, \ours{} outperforms the 
vanilla retriever (Contriever) on the balanced mixture across all NDCG cutoffs 
(e.g., NDCG@1: 47.66 $\rightarrow$ 50.86 with $\lambda$=0.002), confirming 
that decoupled temporal scoring provides clear benefits when temporal 
queries are present, even when mixed with non-temporal ones. On the purely non-temporal subset, \ours{} shows only a marginal drop 
(NDCG@1: 61.47 $\rightarrow$ 60.06 with the default $\lambda$), which is an 
expected behavior given that our TIE module is trained on temporal queries 
only, rather than distinguishing 
temporal queries from non-temporal queries. Yet, this gap can be further reduced 
by tuning $\lambda$, which controls the strength of the temporal score. Overall, these results show that \ours{} remains 
robust under mixed-query settings, highlighting its applicability to 
real-world retrieval.

\begin{table*}[t]
\centering
\scalebox{0.8}{
\begin{tabular}{@{}l|r@{\hspace{10pt}}r@{\hspace{10pt}}r@{\hspace{10pt}}r| r@{\hspace{10pt}}r@{\hspace{10pt}}r@{\hspace{10pt}}r@{}}
\toprule
 & \multicolumn{4}{c|}{\textbf{Simple}} & \multicolumn{4}{c}{\textbf{Complex}} \\
\cmidrule(lr){2-5}\cmidrule(lr){6-9}
\textbf{Method}
& P@1 & P@5 & R@1 & R@5
& P@1 & P@5 & R@1 & R@5 \\
\midrule \midrule
\multicolumn{9}{c}{\textit{Non-temporal and Temporal Retrievers}} \\
\midrule
BM25
& 61.79 & 16.51 & 61.79 & 82.55
& 45.54 & 13.66 & 45.54 & 68.30 \\
Contriever
& 61.32 & 17.55 & 61.32 & 87.74
& 55.95 & 16.46 & 55.95 & 82.29 \\
JinaAI
& 65.57 & 17.64 & 65.57 & 88.21
& \underline{60.42} & 16.99 & \underline{60.42} & 84.97 \\
\midrule
TempRetriever
& 54.72 & 15.00 & 54.72 & 75.00
& 42.86 & 12.62 & 42.86 & 63.10 \\
TsContriever
& \underline{77.36} & \textbf{18.96} & \underline{77.36} & \textbf{94.81}
& 46.43 & 16.07 & 46.43 & 80.36 \\
TSM
& 73.11 & \underline{18.87} & 73.11 & \underline{94.34}
& 57.89 & \underline{17.14} & 57.89 & \underline{85.71} \\
\midrule
\multicolumn{9}{c}{\textit{Temporal Re-rankers}} \\
\midrule
TempRALM \small(+ Contriever)
& 7.55 & 1.89 & 7.55 & 9.43
& 22.92 & 7.80 & 22.92 & 38.99 \\
MRAG \small(+ Contriever)
& 62.26 & 16.23 & 62.26 & 81.13
& 47.02 & 13.75 & 47.02 & 68.75 \\
TimeR4 \small(+ Contriever)
& 59.91 & 16.89 & 59.91 & 84.43
& 41.96 & 14.67 & 41.96 & 73.36 \\
\midrule
TempRALM \small(+ JinaAI)
& 7.08 & 1.98 & 7.08 & 9.91
& 23.96 & 7.98 & 23.96 & 39.88 \\
MRAG \small(+ JinaAI)
& 63.68 & 15.66 & 63.68 & 78.30
& 32.89 & 9.88 & 32.89 & 49.40 \\
TimeR4 \small(+ JinaAI)
& 62.74 & 16.89 & 62.74 & 84.43
& 45.68 & 15.27 & 45.68 & 76.34 \\
\midrule
\rowcolor{gray!15}
\ours{} \small(+ Contriever)
& 75.00 & 18.87 & 75.00 & 94.34
& 59.38 & 16.70 & 59.38 & 83.48 \\
\rowcolor{gray!15}
\ours{} \small(+ JinaAI)
& \textbf{80.19} & 18.87 & \textbf{80.19} & 94.34
& \textbf{65.48} & \textbf{17.23} & \textbf{65.48} & \textbf{86.16} \\
\bottomrule
\end{tabular}}
\vspace{-0.2cm}
\caption{\textbf{Retrieval performance under noisy document timestamps on TempRAGEval.} 
We simulate noisy document timestamps and evaluate all methods on the simple and 
complex subsets. Re-rankers and \ours{} re-rank the top-$k$ documents retrieved 
by Contriever or JinaAI. \textbf{Bold} and \underline{underline} denote the best 
and second-best results.}
\label{tbl:tempragval_noisy}
\end{table*}

\subsection{Robustness to Missing or Noisy Document Time}
\label{supp:noisy_doc_time}
Continuing from \S\ref{subsec:exp_more_analysis}, we further investigate \ours{}'s robustness under timestamp noise, simulating 
a realistic setup where gold document timestamps are unavailable. Instead of 
using gold timestamps, we estimate each document's temporal interval as 
$[\min, \max]$ of the dates mentioned in the document text, extracted via Python regex. 
Table~\ref{tbl:tempragval_noisy} reports the retrieval performance on 
TempRAGEval under this setting.

\ours{} achieves the best performance among temporal re-rankers across both 
subsets, with \ours{}+JinaAI outperforming TempRALM, MRAG, and TimeR4 by 
large margins (e.g., P@1: 80.19 vs.\ at most 63.68 on the simple subset). 
Notably, jointly-trained temporal retrievers (e.g., TsContriever, TSM) appear 
relatively more robust to timestamp noise than re-rankers, likely because 
their learned representations absorb noisy temporal signals more smoothly 
than explicit interval-based scoring. Even so, \ours{} -- though a re-ranker 
itself -- remains competitive with these temporal retrievers on the simple 
subset and surpasses them on the complex subset (e.g., P@1: 65.48 vs.\ TSM's 
57.89), suggesting that decoupled temporal scoring continues to provide value 
even when document timestamps are unreliable. This indicates that \ours{} can be applied to corpora with imperfect 
temporal metadata, and its robustness under heavy timestamp noise could 
be further strengthened by integrating document-side timestamp cleaning 
or estimation techniques~\citep{leeuwenberg2018temporal}.

\begin{table*}[t]
\centering
\scalebox{0.69}{
\begin{tabular}{@{}l|r@{\hspace{8pt}}r@{\hspace{8pt}}r@{\hspace{8pt}}r| r@{\hspace{8pt}}r@{\hspace{8pt}}r@{\hspace{8pt}}r| r@{\hspace{8pt}}r@{\hspace{8pt}}r@{\hspace{8pt}}r| r@{\hspace{8pt}}r@{\hspace{8pt}}r@{\hspace{8pt}}r@{\hspace{4pt}}|}
\toprule
 & \multicolumn{8}{c|}{\textbf{Nobel Prize}} & \multicolumn{8}{c|}{\textbf{TempRAGEval}} \\
 & \multicolumn{8}{c|}{Simple\hspace{100pt}Complex}
 & \multicolumn{8}{c|}{Simple\hspace{100pt}Complex} \\
\cmidrule(lr){2-5}\cmidrule(lr){6-9}
\cmidrule(lr){10-13}\cmidrule(lr){14-17}
\textbf{Method}
& P@1 & P@5 & R@1 & R@5
& P@1 & P@5 & R@1 & R@5
& P@1 & P@5 & R@1 & R@5
& P@1 & P@5 & R@1 & R@5 \\
\midrule \midrule
\multicolumn{17}{c}{\textit{Non-temporal Retriever + Temporal Re-rankers}} \\
\midrule
TempRALM \small(+ Contriever)
& 0.43 & 0.44 & 0.10 & 0.59
& 25.24 & 17.93 & 7.56 & 22.00
& 5.66 & 2.08 & 5.66 & 10.38
& 26.04 & 8.51 & 26.04 & 42.56 \\
MRAG \small(+ Contriever)
& 51.82 & 34.17 & 13.90 & 43.74
& 14.66 & 17.31 & 3.77 & 20.30
& 62.26 & 16.23 & 62.26 & 81.13
& 47.02 & 13.75 & 47.02 & 68.75 \\
TimeR4 \small(+ Contriever)
& 17.69 & 13.14 & 4.72 & 16.89
& 17.31 & 15.00 & 5.82 & 19.73
& 59.91 & 16.98 & 59.91 & 84.91
& 48.96 & 15.33 & 48.96 & 76.64 \\
\rowcolor{gray!15}
\ours{} \small(+ Contriever)
& 60.54 & 45.25 & 17.96 & 58.91
& 31.25 & 29.18 & 11.83 & 47.61
& 88.21 & 19.25 & 88.21 & 96.23
& 79.32 & 17.68 & 79.32 & 88.39 \\
\midrule
TempRALM \small(+ JinaAI)
& 0.71 & 0.56 & 0.25 & 0.85
& 32.45 & 22.84 & 11.94 & 31.32
& 8.02 & 2.08 & 8.02 & 10.38
& 34.08 & 8.72 & 34.08 & 43.60 \\
MRAG \small(+ JinaAI)
& 66.89 & 38.10 & 20.65 & 53.03
& 18.03 & 18.08 & 4.67 & 25.59
& 63.68 & 15.66 & 63.68 & 78.30
& 32.89 & 9.88 & 32.89 & 49.40 \\
TimeR4 \small(+ JinaAI)
& 38.47 & 24.68 & 12.77 & 35.58
& 21.39 & 18.22 & 8.65 & 30.04
& 65.57 & 16.98 & 65.57 & 84.91
& 57.74 & 16.01 & 57.74 & 80.06 \\
\rowcolor{gray!15}
\ours{} \small(+ JinaAI)
& \textbf{78.76} & \textbf{55.96} & \textbf{25.47} & \textbf{73.41}
& \underline{46.88} & \textbf{35.82} & \underline{18.35} & \textbf{62.30}
& \textbf{93.87} & \underline{19.53} & \textbf{93.87} & \underline{97.64}
& \textbf{84.08} & \underline{18.12} & \textbf{84.08} & \underline{90.62} \\
\midrule
\multicolumn{17}{c}{\textit{Temporal Retriever + Temporal Re-rankers}} \\
\midrule
TempRALM \small(+ TSM)
& 0.99 & 0.89 & 0.40 & 1.45
& 29.09 & 13.37 & 10.46 & 19.48
& 7.08 & 2.08 & 7.08 & 10.38
& 28.87 & 8.63 & 28.87 & 43.15 \\
MRAG \small(+ TSM)
& 51.88 & 31.23 & 15.02 & 41.43
& 21.63 & 20.10 & 6.30 & 28.64
& 71.23 & 17.83 & 71.23 & 89.15
& 50.60 & 15.18 & 50.60 & 75.89 \\
TimeR4 \small(+ TSM)
& 38.13 & 27.17 & 10.69 & 35.73
& 23.56 & 17.12 & 8.80 & 26.83
& 71.23 & 18.30 & 71.23 & 91.51
& 54.02 & 16.37 & 54.02 & 81.85 \\
\rowcolor{gray!15}
\ours{} \small(+ TSM)
& \underline{72.50} & \underline{53.21} & \underline{22.25} & \underline{68.06}
& \textbf{51.68} & \underline{32.64} & \textbf{22.25} & \underline{53.74}
& \underline{91.51} & \textbf{19.72} & \underline{91.51} & \textbf{98.58}
& \underline{80.95} & \textbf{18.30} & \underline{80.95} & \textbf{91.52} \\
\bottomrule
\end{tabular}}
\vspace{-0.2cm}
\caption{\textbf{Overall TIR performance with temporal retriever + temporal re-ranker baselines.} We report Precision@$\{1,5\}$ and Recall@$\{1,5\}$ on the simple and complex subsets of Nobel Prize and TempRAGEval datasets. We compare temporal re-rankers (TempRALM, MRAG, TimeR4) and \ours{}, where all re-rankers re-rank the top-$k$ documents retrieved by Contriever, JinaAI, or the best performing temporal retriever TSM in Table~\ref{tbl:nobel_timeqa_retrieval}. \textbf{Bold} and \underline{underline} denote the best and second-best results in each column.}
\label{tbl:more_tir}
\end{table*}

\subsection{Statistical Significance Tests}
\label{supp:stat_sig}

Continuing from \S\ref{subsec:exp_more_analysis}, we detail the statistical
significance tests of \ours{}'s improvements. Since our pipeline is
deterministic at inference (fixed splits, temperature-0 decoding), the
relevant variance lies across queries rather than across runs. We thus
follow standard IR practice and perform paired
per-query tests between \ours{} and all baselines over the identical query
set on TempRAGEval, an OOD setting used in our experiments. Specifically, we use exact McNemar's tests for binary
metrics (e.g., P@1) and two-sided Wilcoxon signed-rank tests for graded
metrics (e.g., NDCG@$k$).

We conduct 270 comparisons in total, spanning 2 base retrievers (Contriever,
JinaAI) $\times$ 9 baselines $\times$ 5 metrics $\times$ 3 query subsets.
\ours{}'s improvements are statistically significant in 246 out of 270
comparisons ($p < 0.05$), and every P@1 and NDCG@5 improvement is
significant at $p < 0.001$, confirming that the gains reported in
Table~\ref{tbl:nobel_timeqa_retrieval} are robust rather than artifacts of
query-level variance.

\subsection{Comparison with More TIR Baselines}
\label{supp:exp_more_tir}

Table~\ref{tbl:more_tir} reports retrieval performance when temporal
re-rankers are paired with the temporal retriever TSM, extending our
analysis to additional TIR baselines. \ours{} remains the strongest
re-ranker regardless of the underlying retriever: paired with TSM,
\ours{} outperforms the strongest competing re-ranker by 20.62\%p
Precision@1 (72.50 vs.\ MRAG+TSM's 51.88) on Nobel Prize simple
queries and by 22.59\%p Precision@1 (51.68 vs.\ TempRALM+TSM's 29.09)
on Nobel Prize complex queries, with similar margins on TempRAGEval.
In addition, while replacing a non-temporal retriever with TSM
benefits all baseline re-rankers, the gains \emph{plateau} for
\ours{}, which already achieves near-saturated performance with
JinaAI (e.g., 93.87 vs.\ 91.51 Precision@1 on TempRAGEval simple
queries). This suggests that \ours{}'s temporal reasoning already
compensates for the lack of temporal awareness in the retriever.

\end{document}